%% file: tmi.tex
\documentclass[journal,twoside,web]{ieeecolor}
\usepackage{tmi}
\usepackage{cite}
\usepackage{amsmath,amssymb,amsfonts}
\usepackage{graphicx}
\usepackage{textcomp}
\usepackage[ruled]{algorithm}
\usepackage{xcolor}
\usepackage{adjustbox}
\usepackage{algpseudocode}
\usepackage{subcaption}
\usepackage{multirow}
\usepackage[colorlinks=true, linkcolor=blue, urlcolor=blue]{hyperref}

\def\BibTeX{{\rm B\kern-.05em{\sc i\kern-.025em b}\kern-.08em
    T\kern-.1667em\lower.7ex\hbox{E}\kern-.125emX}}
\begin{document}
\title{BSDA: Bayesian Random Semantic Data Augmentation for Medical Image Classification}
\author{Yaoyao Zhu, Xiuding Cai, Xueyao Wang, Xiaoqing Chen, Yu Yao, and Zhongliang Fu
\thanks{This work was supported in part by the National Natural Science Foundation of China under Grant 82073338. (Corresponding author: Zhongliang Fu)}
\thanks{Yaoyao Zhu, Xiuding Cai, Xueyao Wang, Xiaoqing Chen, Yu Yao, and Zhongliang Fu are with Chengdu Institute of Computer Application, Chinese Academy of Sciences, Chengdu, China, and the School of Computer Science and Technology, University of Chinese Academy of Sciences, Beijing, China 
(e-mail:zhuyaoyao19@mails.ucas.ac.cn; caidiuding20@gmail.com; wangxueyao221@mails.ucas.ac.cn; chenxiaoqing@casit.com.cn; Casitmed2022@163.com; fzliang@casit.com.cn).}}
\maketitle

\begin{abstract}
    Data augmentation is a crucial regularization technique for deep neural networks, particularly in medical image classification. 
    Mainstream data augmentation (DA) methods are usually applied at the image level. 
    Due to the specificity and diversity of medical imaging, expertise is often required to design effective DA strategies, and improper augmentation operations can degrade model performance. 
    Although automatic augmentation methods exist, they are computationally intensive. 
    Semantic data augmentation can implemented by translating features in feature space. However, over-translation may violate the image label.
    To address these issues, we propose \emph{Bayesian Random Semantic Data Augmentation} (BSDA), a computationally efficient and handcraft-free feature-level DA method. 
    BSDA uses variational Bayesian to estimate the distribution of the augmentable magnitudes,
    and then a sample from this distribution is added to the original features to perform semantic data augmentation.
    We performed experiments on nine 2D and five 3D medical image datasets.
    Experimental results show that BSDA outperforms current DA methods.
    Additionally, BSDA can be easily assembled into CNNs or Transformers as a plug-and-play module, improving the network's performance.
    The code is available online at \url{https://github.com/YaoyaoZhu19/BSDA}.
\end{abstract}


\input{context/intro}

\input{context/rel_work}

\input{context/method}

\input{context/exp}

\input{context/dis}



\bibliographystyle{IEEEtran}
\bibliography{references}

\end{document}

%% file: context/intro.tex
\section{Introduction}
Deep learning methods can assist clinicians in rapid examination and accurate diagnosis~\cite{goceri_medical_2023}.
However, these methods are data-demanding, and medical images are often scarce. 
For example, insufficient patients with specific diseases or a lack of medical equipment can lead to biased, overfitting, and inaccurate models~\cite{goceri_medical_2023}.
Data augmentation is a common regularization technique to address these issues~\cite{Liu2023MedAugmentUA, Medaug}. 
The variety of medical image modalities (e.g., MR, CT, X-ray, Retina) and their association with different clinical diseases necessitate specialized knowledge and careful debugging for image-level data augmentation methods to improve network performance significantly. 
Although automated data augmentation methods~\cite{AutoAugment, Randaugment} can avoid manual debugging, they are computationally expensive. 
Additionally, image-level data augmentation methods often need help to improve sample diversity and achieve semantic transformations. 
While generative data augmentation methods~\cite{chai2022synthetic,li2021semantic, moghadam2023morphology, pinaya2022brain} can enhance semantic diversity, they are also computationally expensive and complex to train, making them challenging for medical practitioners without a computer science background.

\begin{figure}[htbp]
  \centering
  \begin{subfigure}[b]{0.45\textwidth}
    \centering
    \includegraphics[width=\textwidth]{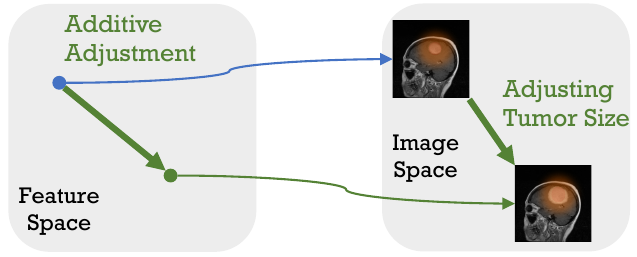}
    \caption{Semantic Data Augmentation}
    \label{fig: semantic_magn}
\end{subfigure}
\vfill
  \begin{subfigure}[b]{0.45\textwidth}
      \centering
      \includegraphics[width=\textwidth]{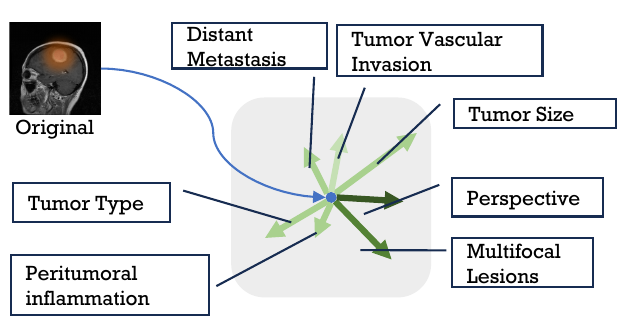}
      \caption{Semantic Directions}
      \label{fig: semantic_dir}
  \end{subfigure}
    \caption{
      Example of semantic data augmentation. 
      \ref{fig: semantic_dir} is a non-exact example of different semantic directions in the feature space along which moving can change the semantics.
      subigure~\ref{fig: semantic_magn} shows a non-exact example of semantic data augmentation in feature space
  }
  \label{fig: semantic_aug_example}
\end{figure}

Recent studies have shown that feature-level data augmentation methods can enhance network performance~\cite{8979439,kang_improving_2023,9711259,10.1145/3503161.3547866}.
In detail, the deep feature space harbors various semantic directions, and translating features along these directions yields new sample features with identical class identities but alter semantic content~\cite{isda_tpami}. 
An example is shown in Figure~\ref{fig: semantic_aug_example}, subfigure~\ref{fig: semantic_dir} shows different semantic directions in feature space of original image input to the deep network, 
and subfigure~\ref{fig: semantic_magn} shown translate deep feature along `tumor size,` we will then obtain the depth features of another image where the tumor size has changed but maintained other semantics.

The ISDA~\cite{isda_tpami} stands out in this domain, facilitating implicit data augmentation by aiming to minimize an upper bound of the expected cross-entropy loss on the augmented dataset.
Unlike traditional methods that modify images directly, this approach generates new data at the feature level, such as operating random disturbances, interpolations, or extrapolations within the feature space for augmentation~\cite{devries2017dataset}.
In medical images, there are numerous modalities with non-uniform dimensions. 
data augmentation methods based on image transformations provide limited enhancement of diversity,
and  computational cost of generative data augmentation methods is high,
the generative data augmentation methods are computationally costly, although they can enrich diversity
Semantic data augmentation holds promise in addressing the shortcomings of both approaches. 
Current semantic data augmentation methods~\cite{isda_tpami, sfa} mainly aim to improve the semantic direction, whereas semantic strength should be equally important because over-translation may violate the image label. 

To address this issue and develop a universal semantic augmentation paradigm for medical image classification, we drew inspiration from the automatic augmentation method: RA~\cite{Randaugment}, defining a semantic data augmentation strategy incorporating two hyperparameters: semantic magnitude and semantic direction. 
Inspired by the concept that modeling data augmentation as additive perturbation can enhance network learning and generalization capabilities~\cite{NEURIPS2022_2130b8a4}, we define a semantic data augmentation method as the addition of semantic magnitude to the original feature in selected semantic directions. 
For example, in the tumor staging task, a feature represents the semantics of tumor size, and we change the tumor size by changing the feature value.

\begin{figure}[t]
  \centering
  \includegraphics[width=0.40\textwidth]{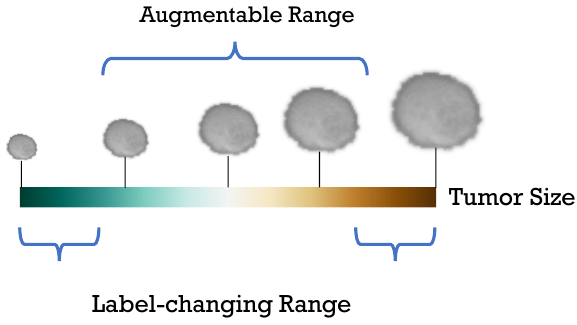}
  \caption{
    Imprecisely example of augmentable semantic magnitude.
    For instance, in the tumor grading task, 
    if the centermost value (white) is an input sample, 
    the curly brackets indicate the range that does not change the original label, 
    and the augmented samples beyond that range change the original label.
      }
      \label{fig: tumor_size}
\end{figure}

However, Figure~\ref{fig: tumor_size} shows that alterations beyond the permissible range within the category may result in label changes.
Therefore, \emph{we treat the augmentable (label-preserving) semantic magnitude as a random variable} and estimate its distribution using variational Bayesian.
For semantic directions, similar to the image space augmentation approache~\cite{he2016deep}, it is a naive idea to select semantic directions randomly, 
but it does not make sense to perform augmentation in certain directions~\cite{isda_tpami}, a view based on image space.
Some data augmentation approaches~\cite{AugMix, CutMix} are adequate for downstream tasks, although they do not make sense vision, 
and the quantitative evaluation approaches proposed in~\cite{yang2024investigating} explain why vision meaningless data augmentation approaches are still practical.
Not coincidentally,~\cite{9711259} point out that adding a random Gaussian perturbation to the features significantly improves the Empirical Risk Minimization (ERM), although it does not follow any meaningful direction.
Even the perturbation of randomness due to the reparameterization introduced by variational inference benefits the network's learning of features~\cite{9711259}.
Therefore, we do not augment all directions but randomly select semantic directions like the random selection transform in image space.

In summary, we propose a simple yet efficient feature-level method for semantic data augmentation called Bayesian Random Semantic Data Augmentation (BSDA). 
BSDA adds semantic magnitudes to randomly selected directions, with the magnitudes obtained by sampling from augmentable semantic magnitude distribution estimated using variational inference.
Figure~\ref{figures: overview} shows BSDA inserted into the network as a plug-in.
Our main contributions are:
\begin{itemize}
  \item We propose a high-performance Bayesian random semantic data augmentation plug-and-play module, BSDA, for medical image classification.
  \item We experimentally demonstrate that BSDA outperforms current data augmentation methods.
  \item We provide experimental evidence demonstrating that BSDA can enhance network performance across \emph{different dimensions, modalities, and neural network architectures}, including both CNNs and Transformers.
\end{itemize}

%% file: context/rel_work.tex
\section{Related Work}
\subsection{Data Augmentation}
In image recognition tasks, data augmentation techniques like random flipping, panning, and rotation make the network learn certain invariants~\cite{he2016deep,tan2019efficientnet}. 
These methods rely heavily on empirical knowledge, and a specific augmentation strategy can only be used for a particular dataset.
AutoAugment~\cite{AutoAugment} was the first technique to perform automatic augmentation by searching for a better strategy across numerous solution spaces using reinforcement learning.
However, due to the large of its search space, other works (e.g., RA~\cite{Randaugment}, DADA~\cite{li2020differentiable}) have also shown strong performance by improving the search algorithm. 
While automated data augmentation algorithms have shown strong capabilities, the training of deep neural networks remains computationally intensive.
Image Erasure~\cite{zhong2020random} and Image Mixing~\cite{zhang2017mixup, AugMix, CutMix} enhance the performance of the network by performing random erasure or mixing of images in the image space. 
Generative models have also shown strong performance in image generation, with some works~\cite{ganaug,diffusionaug} developing data augmentation techniques based on image generation. 
However, these methods require additional training of a generative model for each dataset.
A Bayesian data augmentation method proposed in~\cite{tran2017bayesian} learns the distribution of features from the training set by generating an adversarial network implementing generalized Monte Carlo expectation maximization and samples from it to obtain augmented data.
It is worth noting that our method is similar in name but differs in approach. 
Several studies have explored the efficiency of data augmentation. \cite{NEURIPS2022_2130b8a4} notes that data augmentation modeled as additive perturbation improves network performance by amplifying and perturbing the singular values of the network's Jacobian determinant.
Similarly, \cite{rajput2019does} showed a lower bound on the data required for data augmentation, demonstrating that the general use of DA requires an exponential amount of data before it leads to larger bounds.
\subsection{Semantic Data Augmentation}
Semantic data augmentation is achieved by label-preserving translation in the semantic space, generating new sample points in the feature space.
Three semantic augmentation schemes involve adding random Gaussian noise with mean 0 and variance $\sigma^2$ to the features, interpolating neighbors, and extrapolating~\cite{devries2017dataset}.
Similarly, OnlineAugment~\cite{OnlineAugment} achieves excellent performance by using Gaussian noise to perturb features, but it tends to generate non-class-preserving sample points.
Furthermore, \cite{sfa} used a moving average to estimate the feature covariance matrix online during training and modeled the cross-feature joint noise distribution.
ISDA~\cite{isda_tpami} is a novel semantic data augmentation algorithm that enhances dataset diversity by translating training samples in the deep feature space along semantically meaningful directions, improving the generalization performance of deep models with minimal computational cost.
A shape space-based feature augmentation method projects multiple image features into a pre-shape space.
\cite{pmlr-v202-zhu23i} uses Wasserstein geodesic interpolation to augment data and regularize performance.
Moment Exchange is an implicit data augmentation method that enhances recognition models by replacing and interpolating learned features' moments (mean and standard deviation) between training images.
This forces models to extract training signals from these moments, thus improving generalization across benchmark datasets~\cite{9578208}.

%% file: context/method.tex
\section{Method}
Consideraing training a deep neural network $G_{\Theta} = f_1 \circ f_2$ included two parts: a feature extraction network $f_1$ and a classification network $f_2$ with parameters $\Theta$ on a dataset $\mathcal{D} = \{(\mathbf{x_i}, \mathbf{y_i})\}_{i=1}^N$, where each $\mathbf{y}_i$ represents a label belonging to one of $c$ classes. 
The output of $f_1$ is a $k$ dimensions feature vector $\mathbf{a} = f_1(\mathbf{x}) \in \mathcal{R}^k$, which is then input into $f_2$ to predict the target $\hat{\mathbf{y}} = f_2(\mathbf{a}) = f_2(f_1(\mathbf{x}))$, where $\hat{\mathbf{y}}$ is the predicted class label. 
We refer to $\mathbf{a}$ as the original feature vector for clarity.

\begin{figure*}[htbp]
  \center
  \includegraphics[width=0.8\textwidth]{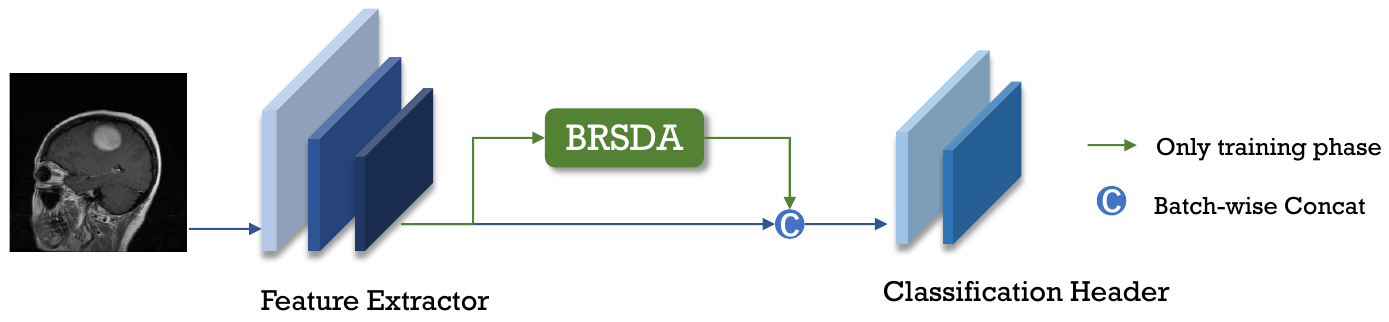}
  \caption{
      Illustrates of BSDA as a plug-in into the nerual network.
      }
      \label{figures: overview}
  \end{figure*}
  
\subsection{BSDA} 
The BSDA method generates augmented feature $\mathbf{\tilde{a}}$ by adding the semantic magnitude $\mathbf{m}$ to the original feature $\mathbf{a}$ 
after element-wise multiplication of them with randomly selected semantic directions $\mathbf{d}_\lambda$. 
The formula is as follows:

\begin{equation}
\mathbf{\tilde{a}} = \mathbf{a} +  \mathbf{d}_\lambda \odot \mathbf{m},
\end{equation}
where $\mathbf{d} \in \left\{0, 1 \right\}^k$ is a binary vector, initially set to all ones and then set to zero with probability $\lambda$, and $\odot$ denotes element-wise multiplication.
$\mathbf{m} \in \mathcal{R}^k$ represents the semantic magnitude sampled from the augmentable magnitude distribution $p(\mathbf{m}|\mathbf{a})$ when given the original feature vector $\mathbf{a}$.
Furthermore, to preserve specific properties of the features, such as low rank~\cite{galanti2022sgd, kamalakara2022exploring, arora2019implicit}, we mask semantic directions corresponding to zero feature values. 
We have:

\begin{equation}
    \mathbf{\tilde{a}} = \mathbf{a} +  \mathbb{I}_{\mathbf{a} \not =0}\mathbf{d}_\lambda \odot \mathbf{m},
    \label{eq:BSDA}
    \end{equation}
where $\mathbb{I}_{\mathbf{a} \not= 0}$ is an indicator function when $\mathbf{a} \not= 0$, the value of this function is 1.

\textbf{Estimate the magnitude distribution.}
To obtain $p(\mathbf{m}|\mathbf{a})$, we introduce a model $q_{\phi_{\mathbf{m}}}(\mathbf{m}|\mathbf{a})$ to approximate the true distribution $p(\mathbf{m}|\mathbf{a})$.
The Kullback-Leibler (KL) divergence measures the similarity between these two distributions, aiming to make $q_{\phi_{\mathbf{m}}}(\mathbf{m}|\mathbf{a})$ closely match $p(\mathbf{m}|\mathbf{a})$ by maximizing the KL divergence.
Thus, our optimization goal is as follows:

\begin{equation}
\tilde{\phi}_{\mathbf{m}} = \underset{\phi_{\mathbf{m}}}{\arg\max} D_{KL}(q_{\phi_{\mathbf{m}}}(\mathbf{m}|\mathbf{a}) || p(\mathbf{m}|\mathbf{a})).
\end{equation}
Removing the terms that are not related to the parameter $\tilde{\phi_{\mathbf{m}}}$ in $D_{KL}$, we have:

\begin{equation}
  \begin{aligned}
& D_{KL}(q_{\phi_{\mathbf{m}}}(\mathbf{m}|\mathbf{a}) || p(\mathbf{m}|\mathbf{a})) \\
&= \text{KL}(q_{\phi_{\mathbf{m}}}(\mathbf{m}|\mathbf{a}) || p(\mathbf{m})) - \mathbb{E}_{\mathbf{m} \sim q_{\phi_{\mathbf{m}}}(\mathbf{m}|\mathbf{a}) }(\log{ p(\mathbf{a} |\mathbf{m}) }).
\label{eq:final_taget}
  \end{aligned}
\end{equation}

\textbf{Loss function of BSDA.}
The first term of Eq.~\ref{eq:final_taget} can be calculated easily. 
The second term estimates the features $\mathbf{a}$ given $\mathbf{m}$, which is more challenging to compute. 
Drawing inspiration from the design of Variational Autoencoders (VAE)~\cite{kingma2013auto}, 
we introduce a reconstruction network for learning this term, rewriting the second term as $p_{\phi_\mathbf{a}}(\mathbf{a}|\mathbf{m})$. 
Then, we obtain the loss function of BSDA as follows:

\begin{equation}
  \begin{aligned}
& \mathcal{L}_{B}(\phi_{\mathbf{m}},\phi_{\mathbf{a}};\mathbf{a}) \\
&= -\text{KL}(q_{\phi_{\mathbf{m}}}(\mathbf{m}|\mathbf{a}) || p(\mathbf{m})) 
    + \mathbb{E}_{\mathbf{m} \sim q_{\phi_{\mathbf{m}}}(\mathbf{m}|\mathbf{a}) }(\log{ p_{\phi_{\mathbf{a}}}(a|m) }).
    \label{eq:BSDA_loss}
  \end{aligned}
\end{equation}
The second part of Eq.~\ref{eq:BSDA_loss} depends on the model, and we use MSE loss in BSDA. 
Assuming the marginal distribution $p(\mathbf{m})$ follows a normal distribution $\mathcal{N}(0, \mathbf{I})$ and $q_{\phi_{\mathbf{m}}}(\mathbf{m}|\mathbf{a})$ also follows a normal distribution $\mathcal{N}(0, \boldsymbol{\sigma}^2)$, 
setting the mean to zero because we aim to learn the offset relative to the original rather than the augmented feature.
Thus, the loss function of BSDA is given by:
\begin{equation}
  \begin{aligned}
& \mathcal{L}_{B}(\phi_{\mathbf{m}},\phi_{\mathbf{a}};\mathbf{a}) = \\
& - \frac{1}{2}\sum_{i=0}^N(1 + \log(\boldsymbol{\sigma}^2) - \boldsymbol{\sigma}^2 ) + 
\frac{1}{2N}\sum_{l=1}^N (\mathbf{\hat{a}} - \mathbf{a})^2,
\label{eq:loss_BSDA_reparam}
\end{aligned}
\end{equation}
where $\boldsymbol{\sigma}^2$ is estimated variance of BSDA, and $\mathbf{\hat{a}}$ is reconstructed feature using $\mathbf{m}$.

\textbf{The reparameterization trick.}
We employ the reparameterization trick to facilitate the computation of the loss function while ensuring the gradient flow for effective backpropagation.
The random variable $\mathbf{m}$ can be represented as a deterministic variable $\mathbf{m}=g_{\phi_{\mathbf{m}}}(\boldsymbol{\epsilon}, \mathbf{a})$, 
with $\boldsymbol{\epsilon} \sim \mathcal{N}(0, \mathbf{I})$ being an auxiliary variable with an independent marginal distribution $p(\boldsymbol{\epsilon})$, 
and $g_{\phi_{\mathbf{m}}}$ being a vector-valued function parameterized by $\phi_{\mathbf{m}}$.

\textbf{Loss function.}
Our augmentation method is training with $G_{\Theta}$. 
For convenience, we denote the loss of $G_{\Theta}$ as $\mathcal{L}_{task}$, which typically uses cross-entropy loss in classification tasks. 
Therefore, the total loss function is:
\begin{equation}
    \mathcal{L} =  \mathcal{L}_{task}^\mathbf{a} + \alpha( \mathcal{L}_{B} + \mathcal{L}^{\tilde{ \mathbf{a}}}_{task}).
    \label{eq:loss}
\end{equation}
Different superscripts on $\mathcal{L}_{task}$ distinguish between augmented and original features. 
The hyperparameter $\alpha$ is a dynamic value introduced to mitigate the impact of BSDA on the network during the initial stages of training when the network has yet to learn valuable features.

To summarize, the BSDA technique can be integrated into deep networks. 
We provide the pseudocode of BSDA in Algorithm~\ref{algorithm:BSDA}.

\begin{algorithm}
  \caption{The BSDA algorithm}
  \begin{algorithmic}[1]
  \State \textbf{Input:} $\mathcal{D}$ ;
  \State Randomly initialize $\Theta$, $\phi_{\mathbf{a}}$, and $\phi_{\mathbf{m}}$ ;
  \For{$t=0$ to $T$}
      \State Sample a mini-batch $\{ \mathbf{x}_i, \mathbf{y}_i \}_{i=1}^B$ from $\mathcal{D}$ ;
      \State Compute features $\mathbf{a}_i = G(\mathbf{x}_i)$ ;
      \State Estimate variance of magnitude distribution $\boldsymbol{\sigma}_i$ ;
      \State Compute \emph{magnitude $\mathbf{m}_i$} using reparameterization trick $\mathbf{m}_i = \boldsymbol{\sigma}_i \odot \boldsymbol{\epsilon}_i$ ;
      \State Compute \emph{augmented feature $\tilde{\mathbf{a}}_i$} according to Eq.~\ref{eq:BSDA} ;
      \State Compute reconstructed feature $\mathbf{\hat{a}}_i$ ;
      \State Compute $\mathcal{L}$ according to Eq.(\ref{eq:loss}) ;
      \State Update $\Theta$, $\phi_{\mathbf{a}}$, and $\phi_{\mathbf{m}}$ ;
  \EndFor
  \end{algorithmic}
  \label{algorithm:BSDA}
\end{algorithm}

\subsection{Compare with other methods}
\textbf{VAE.}
Our approach resembles VAE~\cite{kingma2013auto}, where a latent variable is estimated and used for reconstruction.
However, the relationship between our latent variable $\mathbf{m}$ and the inputs differs from  VAE~\cite{kingma2013auto}, the latent variable representation of the inputs in VAE~\cite{kingma2013auto}.
Our latent variable $\mathbf{m}$ represents the augmentable range in a given feature $\mathbf{a}$ without altering the label.

\textbf{ISDA.}
Our approach differs from ISDA~\cite{isda_tpami} in several key aspects: 
(a) it does not depend on the task's loss function; 
(b) it serves as an implicit data augmentation method, complementing the explicit ISDA. 
This makes BSDA more convenient for special treatments in particular network branches. 
For instance, the BSDA module can be applied to both branches simultaneously in feature-decoupled networks, achieving post-decoupled feature enhancement.

\begin{figure*}[ht]
  \centering
  \includegraphics[width=\textwidth]{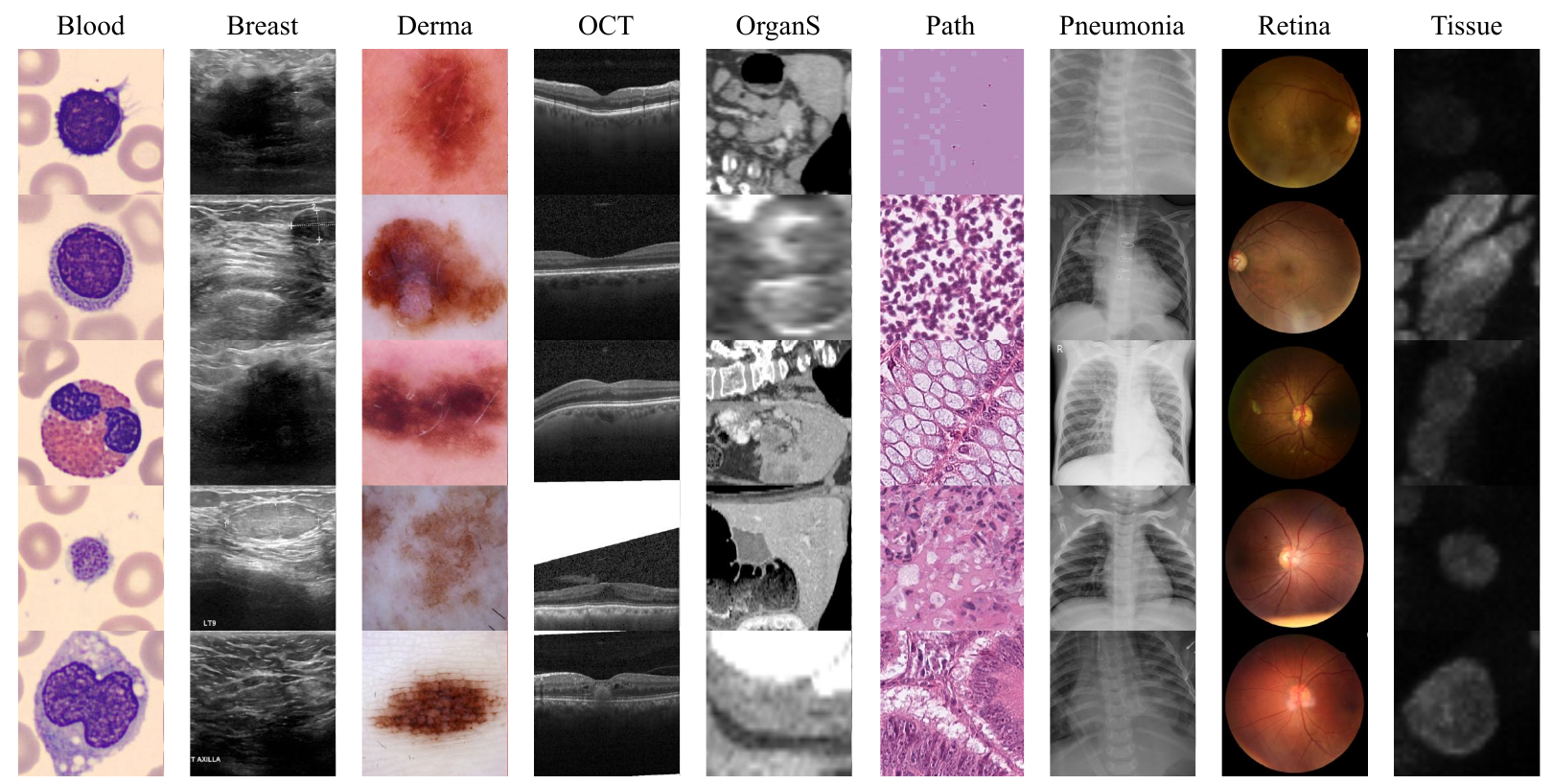}
  \caption{
      Samples of MedMNIST+. 
      For convenience, we removed the MNIST suffix from the dataset. 
      }
      \label{figures: dataset}
  \end{figure*}

%% file: context/exp.tex
\section{Experiments}
In this section, we empirically validate the proposed algorithm on MedMNIST+~\cite{yang_medmnist_2023}, a large-scale collection of standardized biomedical images.
Our evaluation strategy encompassed several vital aspects: comparison with state-of-the-art methods, effectiveness across modalities and dimensions, and adaptability to different neural network architectures.
Finally, we performed ablation experiments, hyperparameter analysis, and visualization of deep features.

\subsection{Datasets and Training Details}

\textbf{Dataset.}
The MedMNIST+~\cite{yang_medmnist_2023} dataset comprises twelve pre-processed 2D datasets and six pre-processed 3D datasets from selected sources covering primary data modalities (e.g., X-ray, OCT, Ultrasound, CT, Electron Microscope), 
diverse classification tasks (binary/multi-class, ordinal regression, and multi-label) and data scales (from 100 to 100,000)~\cite{yang_medmnist_2023}. 
We selected nine 2D medical image datasets and six 3D medical image datasets in MedMNIST+~\cite{yang_medmnist_2023} covering twelve modalities, as shown in Figure~\ref{figures: dataset}.
For more details on the dataset, please refer to Table~\ref{tab:dataset-overview}.

\begin{table*}[h]
  \centering
  \renewcommand\arraystretch{1.2}
  \caption{An Overview of Datasets.
  We selected nine 2D medical image datasets and six 3D medical image datasets in MedMNIST+ covering twelve modalities.
  For convenience, we removed the MNIST suffix from the dataset. 
  }
  \label{tab:dataset-overview}
  \begin{adjustbox}{max width=0.8\linewidth}
  \begin{tabular}{cccccc}
  \hline
  \textbf{Dataset} & \textbf{Data Modality} & \textbf{Tasks (Classes/Labels)} & \textbf{Samples} & \textbf{Training/Validation/Test} \\
   \hline
  Blood & Blood Cell Microscope & Multi-Class (8) & 17,092 & 11,959 / 1,712 / 3,421 \\
  Breast & Breast Ultrasound & Binary-Class (2) & 780 & 546 / 78 / 156 \\
  Derma & Dermatoscope & Multi-Class (7) & 10,015 & 7,007 / 1,003 / 2,005 \\
  OCT & Retinal OCT & Multi-Class (4) & 109,309 & 97,477 / 10,832 / 1,000 \\
  OrganS & Abdominal CT & Multi-Class (11) & 25,211 & 13,932 / 2,452 / 8,827 \\
  Path & Colon Pathology & Multi-Class (9) & 107,180 & 89,996 / 10,004 / 7,180 \\
  Pneumonia & Chest X-Ray & Binary-Class (2) & 5,856 & 4,708 / 524 / 624 \\
  Retina & Fundus Camera & Ordinal Regression (5) & 1,600 & 1,080 / 120 / 400 \\
  Tissue & Kidney Cortex Microscope & Multi-Class (8) & 236,386 & 165,466 / 23,640 / 47,280 \\
  \hline
  Organ3D & Abdominal CT & Multi-Class (11) & 1,742 & 971/161/610 \\
  Nodule3D & Chest CT & Binary-Class (2) & 1,633 & 1,158/165/310 \\
  Adrenal3D & Abdominal CT & Binary-Class (2) & 1,584 & 1,188/98/298 \\
  Fracture3D & Chest CT & Multi-Class (3) & 1,370 & 1,027/103/240 \\
  Vessel3D & Brain MRA & Binary-Class (2) & 1,908 & 1,335/191/382 \\
  \hline
  \end{tabular}
\end{adjustbox}
\end{table*}

\textbf{Evaluation Protocols}
We used the MedMNIST+ split training set and validation set to select hyperparameters and reported the results of the test set.
The Randomness arising from model selection is often ignored.
For instance, does method A outperform method B only because the random search for A got lucky~\cite{domainbed}?
Therefore, we repeated the process three times with different random seeds. 
Every number we report is the mean of these repetitions and their estimated standard error.
The area under the ROC curve (AUC) and accuracy (ACC) are used as the evaluation metrics.

\textbf{Implementation Details.}
We implemented BSDA using PyTorch 2.2.2 and Torchvision 0.17.2 and experimented on an NVIDIA RTX 4090 GPU $\times 2$ with an Intel 14900k CPU. 
During training, we utilized the AdamW~\cite{loshchilov2017decoupled} optimizer with a learning rate of 0.001 and employed a learning rate warm-up strategy for the first five epochs.
The 2D image size is $224 \times 224$, and the 3D image size is $64 \times 64 \times 64$.
To ensure fairness, we maintained consistent training configurations across all experiments. 
The distribution estimator and reconstruction modules of BSDA consist of two fully connected layers, followed by BatchNorm and GeLU activation.

\subsection{Comparison Experiment}
First, we evaluated state-of-the-art methods on nine 2D medical image datasets in MedMNIST2D~\cite{yang_medmnist_2023}(BloodMNIST, BreastMNIST, DermaMNIST, OCTMNIST, OrgansMNIST, PathMNIST, PneumoniaMNIST, RetinaMNIST, and TissueMNIST), which include a total of \emph{513,429} samples.
We compared BSDA with state-of-the-art methods, including \emph{Cutout}~\cite{cutout}, \emph{Mixup}~\cite{zhang2017mixup}, \emph{CutMix}~\cite{CutMix}, and \emph{ISDA}~\cite{isda_tpami}., on nine 2D medical image datasets.

\begin{table*}[htbp]
  \centering
    \renewcommand\arraystretch{1.2}
  \caption{Comparing ACC\% (accuracy) performance with state-of-the-art methods. 
  We report mean values and standard deviations in three independent experiments. 
  The \emph{best} results are \textbf{bold-faced} and \underline{underlined} remarked \emph{second} one.
  For convenience, we removed the MNIST suffix from the dataset. 
  Official is the result reported in the MedMNIST+~\cite{yang_medmnist_2023}, and Baseline is our implementation.
  }
  \label{tab:result-acc-2d}
  \begin{adjustbox}{max width=\linewidth}
  \begin{tabular}{lcccccccccc}
    \hline
    \textbf{Method} & \textbf{Blood} & \textbf{Breast} & \textbf{Derma} & \textbf{OCT} & \textbf{OrganS} & \textbf{Path} & \textbf{Pneumonia} & \textbf{Retina} & \textbf{Tissue} & \textbf{Avg} \\
    \hline
      Official & $95.8$ & $83.3$ & $75.4$ & $76.3$ & $77.8$ & $90.9$ & $86.4$ & $49.3$ & $68.1$ & $78.1 $ \\
      Baseline & $98.5 \pm 0.1$ & $81.0 \pm 8.5$ & $74.9 \pm 1.1$ & $86.7 \pm 1.3$ & $79.5 \pm 0.2$ &  $91.8 \pm 0.6$ & $82.1 \pm 6.6$ & $47.4 \pm 5.5$ & $69.4 \pm 0.9$ & $79.0$ \\
      Mixup & $98.4 \pm 0.2$ & $83.5 \pm 3.2$ & $\underline{76.6 \pm 0.9}$ & $\bf{89.5 \pm 0.8}$ & $80.5 \pm 0.8$ & $89.5 \pm 2.4$ & $81.6 \pm 6.1$ & $51.3 \pm 0.9$ &  $\bf{70.4 \pm 0.5}$ & $80.1$ \\
      Cutout & $98.2 \pm 0.1$ & $\bf{86.3 \pm 3.7}$ & $75.6 \pm 0.1$ & $87.9 \pm 1.9$ & $80.1 \pm 1.4$ & $\bf{92.3 \pm 1.5}$ & $86.1 \pm 0.5$ & $51.5 \pm 4.9$ & $69.5 \pm 0.9$ & $80.8 $ \\
      CutMix & $98.3 \pm 0.1$ & $84.6 \pm 0.6$ & $76.3 \pm 0.5$ & $87.5 \pm 0.3$ & $80.6 \pm 0.5$ & $90.3 \pm 2.3$ & $83.6 \pm 7.5$ & $52.2 \pm 1.5$ & $\underline{69.7 \pm 0.0}$ & $80.4$ \\
      ISDA & $\underline{98.7 \pm 0.4}$ & $\underline{86.1 \pm 1.0}$ & $\bf{76.7 \pm 0.4}$ & $88.2 \pm 1.6$ & $\underline{80.8 \pm 1.3}$ & $90.9 \pm 0.7$ & $\underline{87.2 \pm 3.7}$ &  $\underline{52.6 \pm 1.5}$ & $68.1 \pm 2.6$ &  $\underline{81.0}$ \\
      \textbf{BSDA(Our)} & $\bf{98.8 \pm 0.1}$ & $\underline{86.1 \pm 1.5}$ &  $76.4 \pm 0.8$ & $\underline{88.8 \pm 1.3}$ & $\bf{82.2 \pm 0.7}$ & $\underline{91.9 \pm 3.2}$ & $\bf{88.8 \pm 1.2}$ & $\bf{53.3 \pm 0.1}$ & $\bf{70.4 \pm 1.0}$ & $\bf{81.9}$ \\
    \hline 
  \end{tabular}
\end{adjustbox}
\end{table*}

\begin{table*}[htbp]
  \centering
  \renewcommand\arraystretch{1.2}
\caption{Comparing the AUC\% (area under the ROC curve) performance with state-of-the-art methods. 
We report mean values and standard deviations in three independent experiments. 
The \emph{best} results are \textbf{bold-faced} and \underline{underlined} remarked \emph{second} one.
For convenience, we removed the MNIST suffix from the dataset. 
Official is the result reported in the MedMNIST+~\cite{yang_medmnist_2023}, and Baseline is our implementation.
}
  \label{tab:result-auc-2d}
  \begin{adjustbox}{max width=\linewidth}
    \begin{tabular}{lcccccccccc}
      \hline
      \textbf{Method} & \textbf{Blood} & \textbf{Breast} & \textbf{Derma} & \textbf{OCT} & \textbf{OrganS} & \textbf{Path} & \textbf{Pneumonia} & \textbf{Retina} & \textbf{Tissue} & \textbf{Avg} \\
      \hline
      Official & $99.8$ & $89.1$ & $92.0$ & $95.8$ & $97.4$ & $98.9$ & $95.6$ & $71.0$ & $93.3$ & $92.5$ \\
      Baseline & $\bf{99.9 \pm 0.0}$ & $89.6 \pm 2.0$ & $\bf{93.2 \pm 0.3}$ & $98.7 \pm 0.1$ & $97.7 \pm 0.0$ & $\bf{99.4 \pm 0.1}$ & $95.1 \pm 0.9$ & $72.6 \pm 1.8$ & $93.6 \pm 0.4$ & $93.3$ \\
      Mixup & $\bf{99.9 \pm 0.0}$ & $89.5 \pm 1.2$ & $92.7 \pm 0.5$ & $98.7 \pm 0.3$ & $97.8 \pm 0.2$ & $98.7 \pm 0.4$ & $95.8 \pm 0.4$ & $71.9 \pm 1.3$ & $\bf{93.9 \pm 0.1}$ & $93.2$ \\
      Cutout & $\bf{99.9 \pm 0.0}$ & $\underline{91.1 \pm 1.5}$ & $93.0 \pm 0.5$ & $98.8 \pm 0.6$ & $97.8 \pm 0.2$ & $99.0 \pm 0.5$ & $\underline{95.9 \pm 0.6}$ & $72.5 \pm 1.4$ & $93.6 \pm 0.3$ & $93.5$ \\
      CutMix & $\bf{99.9 \pm 0.0}$ & $90.7 \pm 1.0$ & $92.9 \pm 0.4$ & $\underline{98.9 \pm 0.2}$ & $\underline{97.9 \pm 0.1}$ & $99.0 \pm 0.5$ & $\bf{96.4 \pm 0.6}$ & $73.4 \pm 1.3$ & $93.9 \pm 0.0$ & $\underline{93.7}$ \\
      ISDA & $\bf{99.9 \pm 0.0}$ & $89.3 \pm 2.0$ & $93.0 \pm 0.4$ & $\bf{99.0 \pm 0.3}$ & $97.8 \pm 0.3$ & $\bf{99.4 \pm 0.1}$ & $95.0 \pm 1.1$ & $\underline{74.1 \pm 1.4}$ & $93.6 \pm 0.4$ & $93.5$ \\
      \textbf{BSDA(Our)} & $\bf{99.9 \pm 0.0}$ & $\bf{91.4 \pm 0.2}$ & $\underline{93.1 \pm 0.2}$ & $\underline{98.9 \pm 0.3}$ & $\bf{97.9 \pm 0.0}$ & $\underline{99.2 \pm 0.4}$ & $95.7 \pm 0.2$ & $\bf{75.0 \pm 0.7}$ & $\underline{93.7 \pm 0.4}$ & $\bf{93.9}$ \\
      \hline
  \end{tabular}
\end{adjustbox}
\end{table*}

\subsubsection{ACC Results of MedMNIST2D+}
Table~\ref{tab:result-acc-2d} reveals that BSDA is the top-performing method, achieving the highest average accuracy of 81.9\% across all evaluated datasets. 
ISDA also performs strongly with an average accuracy of 81.0\%, but is slightly less consistent than BSDA.
This demonstrates the advantages of semantic data augmentation methods for medical images and highlights that BSDA outperforms ISDA.
For example, BSDA achieved 70.4\% accuracy on TissueMNIST and 82.2\% on OrgansMNIST, while ISDA achieved 68.1\% and 80.8\% on these datasets, respectively.
Methods like CutMix~\cite{CutMix}, CutOut~\cite{cutout}, and MixUp~\cite{zhang2017mixup} offer comparable results, with average accuracies of 80.4\%, 80.8\%, and 80.1\%, respectively, yet none consistently surpass the performance of ISDA and BSDA.
RetinaMNIST is the most challenging dataset, with all methods showing lower accuracy levels around 50-53\%, such as ISDA at 52.6\% and BSDA at 53.3\%. 
The results highlight the critical role of data augmentation techniques in enhancing model performance, as the baseline consistently underperforms with an average accuracy of 79.0\%, 
emphasizing the necessity of employing sophisticated augmentation strategies in medical imaging tasks.

\subsubsection{AUC Results of MedMNIST2D+}
The results of Table~\ref{tab:result-auc-2d} reinforces the effectiveness of BSDA, 
which not only achieves the highest average AUC but also consistently performs across different datasets.
BSDA achieves an average AUC of 93.9\%, indicating its effectiveness in enhancing the model's ability to distinguish between classes.
This performance is slightly higher than other methods.
BSDA exhibits lower variability in performance across most datasets, indicating more consistent results. 
For instance, its standard deviation in AUC is relatively low, especially compared to methods like Mixup~\cite{zhang2017mixup} and ISDA~\cite{isda_tpami}, which show higher variability in datasets like PneumoniaMNIST and RetinaMNIST.
ISDA and CutMix~\cite{CutMix} also perform well but with higher variability. 
BSDA consistently outperforms other augmentation techniques in average accuracy (ACC) and area under the curve (AUC) across multiple MedMNIST datasets, demonstrating superior performance and reliability in medical image classification tasks.

\subsection{Results of MedMNIST3D+}
We selected five 3D MedMNIST datasets to validate the performance of BSDA.
As shown in Table~\ref{tab:dataset_compare}. 
BSDA enhances the performance of ResNet-18 across various 3D datasets, with improvements in both AUC and ACC for most datasets.
For example, AdernalMNIST3D shows a substantial accuracy improvement from $71.5\%$ to $89.8\%$ and a slight AUC increase from $89.1\%$ to $89.2\%$, suggesting that BSDA significantly enhances the model's performance, particularly in accuracy. 
OrganMNIST3D shows an accuracy increase from $87.2\%$ to $88.7\%$ and an AUC improvement from $99.2\%$ to $99.4\%$. 
This indicates that BSDA improves the correct classification rate and enhances the model's ability to distinguish between classes. 
Similarly, NoduleMNIST3D experiences a rise in accuracy from $84.5\%$ to $86.1\%$ and in AUC from $87.9\%$ to $89.2\%$, 
showing consistent improvements in both metrics. 
For FractureMNIST3D, accuracy rises from $53.5\%$ to $55.0\%$, 
while AUC improves from $72.7\%$ to $74.0\%$. 
These moderate improvements indicate that BSDA helps in better identifying fractures, 
though the dataset remains challenging. 
VesselMNIST3D also benefits, with accuracy increasing from $92.1\%$ to $93.2\%$ and AUC from $90.2\%$ to $91.7\%$, 
reflecting improved precision and robustness in vessel identification. 
These results underscore the value of data augmentation techniques in improving both the discriminative power and the accuracy of deep learning models in medical imaging tasks, 
demonstrating BSDA's effectiveness in enhancing model performance across diverse medical datasets.

\begin{table*}[htbp]
  \caption{
      Performance of different 3D datasets on the test set using ResNet-18~\cite{he2016deep}. 
      The \emph{better} result is \textbf{bold-faced} compared with the Baseline. 
      Official is the result reported in the MedMNIST+~\cite{yang_medmnist_2023}, and Baseline is our implementation.
      For convenience, we removed the MNIST suffix from the dataset. 
  }
    \renewcommand\arraystretch{1.2}
  \label{tab:dataset_compare}
  \centering
  \begin{tabular}{lcccccc}
  \hline
  \multirow{2}{*}{\textbf{Dataset}} & \multicolumn{3}{c}{ACC\%} & \multicolumn{3}{c}{AUC\%}  \\ 
  \cline{2-7}
                & Official  & Baseline & \textbf{BSDA(Our)} & Official  & Baseline & \textbf{BSDA(Our)}  \\
  \hline
  Organ3D   & $90.7$ & $87.2 \pm 0.7$ & $\bf{88.7 \pm 1.4}$ & $99.6$ & $99.2 \pm 0.1$ & $\bf{99.4 \pm 0.1}$ \\
  Nodule3D  & $84.4$ & $84.5 \pm 1.1$ & $ \bf{86.1 \pm 0.6} $ & $86.3$ & $87.9 \pm 1.5$ & $ \bf{89.2 \pm 1.3}$ \\
  Adernal3D & $72.1$ & $71.5 \pm 2.3$ & $ \bf{83.8} \pm 3.6$ & $82.7$ & $89.1 \pm 2.7$ & $ \bf{89.2} \pm 3.4 $ \\
  Fracture3D & $50.8$ & $53.5 \pm 1.3$ & $\bf{56.9 \pm 4.6}$ & $71.2$ & $72.7 \pm 0.5$ & $\bf{73.1 \pm 4.3}$ \\
  Vessel3D   & $87.7$ & $92.1 \pm 1.1$ & $\bf{93.2 \pm 0.3}$ & $87.4$ & $90.2 \pm 5.5$ & $\bf{91.7 \pm 5.8}$ \\
  \hline
\end{tabular}
\end{table*}

\begin{table*}[htbp]
  \centering
  \caption{
      Evaluation of BSDA on different convolutional neural networks using the test set of PneumoniaMNIST~\cite{yang_medmnist_2023}. 
      The \emph{best} results are \textbf{bold-faced} while the number in brackets denotes the performance improvements achieved by BSDA. 
      The last column is about the additional time introduced by BSDA.}
    \renewcommand\arraystretch{1.2}
  \label{tab:network-comparison}
  \begin{tabular}{lccccc}
  \hline
  \multirow{2}{*}{Network}  & \multicolumn{2}{c}{ACC \%} & \multicolumn{2}{c}{AUC \%} & \multirow{2}{*}{Additional Time \%} \\
  \cline{2-5}
                  & Baseline & {\bfseries BSDA}  & Baseline & {\bfseries BSDA}  & \\
  \hline
  ResNet-18       & $82.1 \pm 6.6$ & $\bf{88.8 \pm 1.2}$ & $95.1 \pm 0.9$ & $\bf{95.7 \pm 0.2}$ & $3.7\%$ \\
  ResNet-50       & $87.0 \pm 3.5$ & $86.3 \pm 3.4$ & $96.8 \pm 0.7$ & $\bf{96.9 \pm 0.2}$ & $5.9\%$ \\
  EfficientNet-B0  & $84.2 \pm 1.0$ & $\bf{84.4 \pm 0.3}$ & $94.3 \pm 0.6$ & $\bf{94.9 \pm 0.5}$ &  $7.3\%$\\
  DenseNet-121     & $84.9 \pm 3.3$ & $\bf{89.4 \pm 2.6}$ & $96.6 \pm 0.3$ & $\bf{96.9 \pm 0.9}$ & $1.5\%$ \\
  ViT-T        & $82.9 \pm 4.4$ & $\bf{86.0 \pm 1.7}$ & $94.9 \pm 0.1$ & $\bf{96.0 \pm 0.6}$ & $7.5\%$ \\
  ViT-S        & $81.1 \pm 5.9$ & $\bf{87.2 \pm 1.8}$ & $95.3 \pm 0.6$ & $\bf{95.9 \pm 0.6}$ & $5.8\%$ \\
  ViT-B        & $81.8 \pm 2.2$ & $\bf{86.8 \pm 0.9}$ & $94.1 \pm 1.0$ & $\bf{95.2 \pm 0.9}$ & $2.3\%$ \\
  Swin-T        & $73.6 \pm 11.4$ & $\bf{77.0 \pm 4.1}$ & $87.3 \pm 4.7$ & $\bf{92.0 \pm 0.2}$ & $1.4\%$ \\
  Swin-S        & $63.9 \pm 1.3$ & $\bf{71.7 \pm 6.6}$ & $81.9 \pm 9.4$ & $\bf{90.6 \pm 1.1}$ & $2.1\%$ \\
  Swin-B        & $62.5 \pm 0.0$ & $62.5 \pm 0.0$ & $88.3 \pm 1.9$ & $88.3 \pm 2.5$ & $1.3\%$ \\
  \hline
  \end{tabular}
\end{table*}

\subsection{Applicability of BSDA}
\label{sec:add_net}
We selected several mainstream architectures in computer vision to conduct experiments to verify the performance of BSDA combined with different models, including ResNet~\cite{he2016deep}, EfficientNet~\cite{tan2019efficientnet}, DenseNet~\cite{huang2017densely}, TinyViT~\cite{wu2022tinyvit}, ViT~\cite{vit}, Swin Transformer~\cite{liu2021swin}.
Table~\ref{tab:network-comparison} shows that the BSDA technique significantly improves accuracy (ACC) and area under the curve (AUC) across different convolutional neural networks, 
particularly on the DenseNet-121, ViT, and Swin Transformer series.
For example, the accuracy of DenseNet-121 improved from 84.9\% to 89.4\%, and the AUC increased from 96.6\% to 96.9\%.
In most cases, BSDA introduces a small additional time cost, up to 7.3\% for EfficientNet-B0.
Although BSDA leads to slight performance degradation on some networks (e.g., ResNet-50), overall, it improves model performance with manageable additional computational costs.
Therefore, BSDA is a technique worth considering for improving model performance.

\subsection{Ablation Study}
\begin{table}[t]
  \centering
  \caption{
      Ablation Study on OragnSMNIST. $-$ means to remove the corresponding module and $+$ means add module. Indicator is indicator function in Eq~\ref{eq:BSDA}, Recon is reconstructor.
  }
    \renewcommand\arraystretch{1.2}
  \label{tab: ablation_study}
  \begin{tabular}{lccccc}
    \hline
    \textbf{Setting} & \textbf{ACC\%} & \textbf{AUC\%} \\
  \hline
  Base & $79.5 \pm 0.2$ & $97.7 \pm 0.0$  \\
  \hline
  BSDA & $82.2 \pm 0.7$ & $97.9 \pm 0.0$  \\
  ~~$-$ Indicator & $77.4 \pm 0.7$ & $97.7 \pm 0.2$  \\
  ~~$-$ Recon & $76.8 \pm 1.8$ &$97.6 \pm  0.2$ \\
  ~~$-$ Indicator and Recon & $80.2 \pm 0.4$ & $97.8 \pm 0.1$  \\
  Random Noise & $79.6 \pm 0.3$ & $97.9 \pm 0.1$ \\
  ~~$+$ Indicator & $79.3 \pm 0.3$ & $97.8 \pm 0.1$ \\
  \hline
  \end{tabular}
\end{table}

We analyzed the effectiveness of BSDA components and compared BSDA by adding random standard Gaussian noise to the latent space features.
Table~\ref{tab: ablation_study} indicates that random noise can somewhat improve model performance, but the improvement is limited.
The BSDA model, with the introduction of the indicator function (Eq.~\ref{eq:BSDA}) and reconstructor components, shows significant improvement over the baseline model, especially in terms of accuracy.
Both the indicator function and reconstructor components are essential for BSDA. 
Notably, model performance degrades considerably when the indicator function and reconstructor components are removed. 
This is equivalent to eliminating the reconstructor as well as the second term in Eq.~\ref{eq:BSDA_loss}, which is comparable to simply making $q_{\phi_{\mathbf{m}}}(\mathbf{m}|\mathbf{a}) $ similar to the standard Gaussian distribution. 
The model performance should be comparable to that of sampling noise from a standard Gaussian distribution added to the original features, 
because it is simultaneously constrained by the loss of the task, thus further making the samples sampled by the model $q_{\phi_{\mathbf{m}}}(\mathbf{m}|\mathbf{a})$ more favorable to the task.

\subsection{Sensitivity Analysis}
To better understand the BSDA method, we conducted a series of experiments analyzing it from different perspectives.
The following sections provide a detailed analysis of the results, including hyperparameter sensitivity analysis and loss weight analysis.

\begin{figure}[!t]
  \centering
  \begin{subfigure}[b]{0.24\textwidth}
      \centering
      \includegraphics[width=\textwidth]{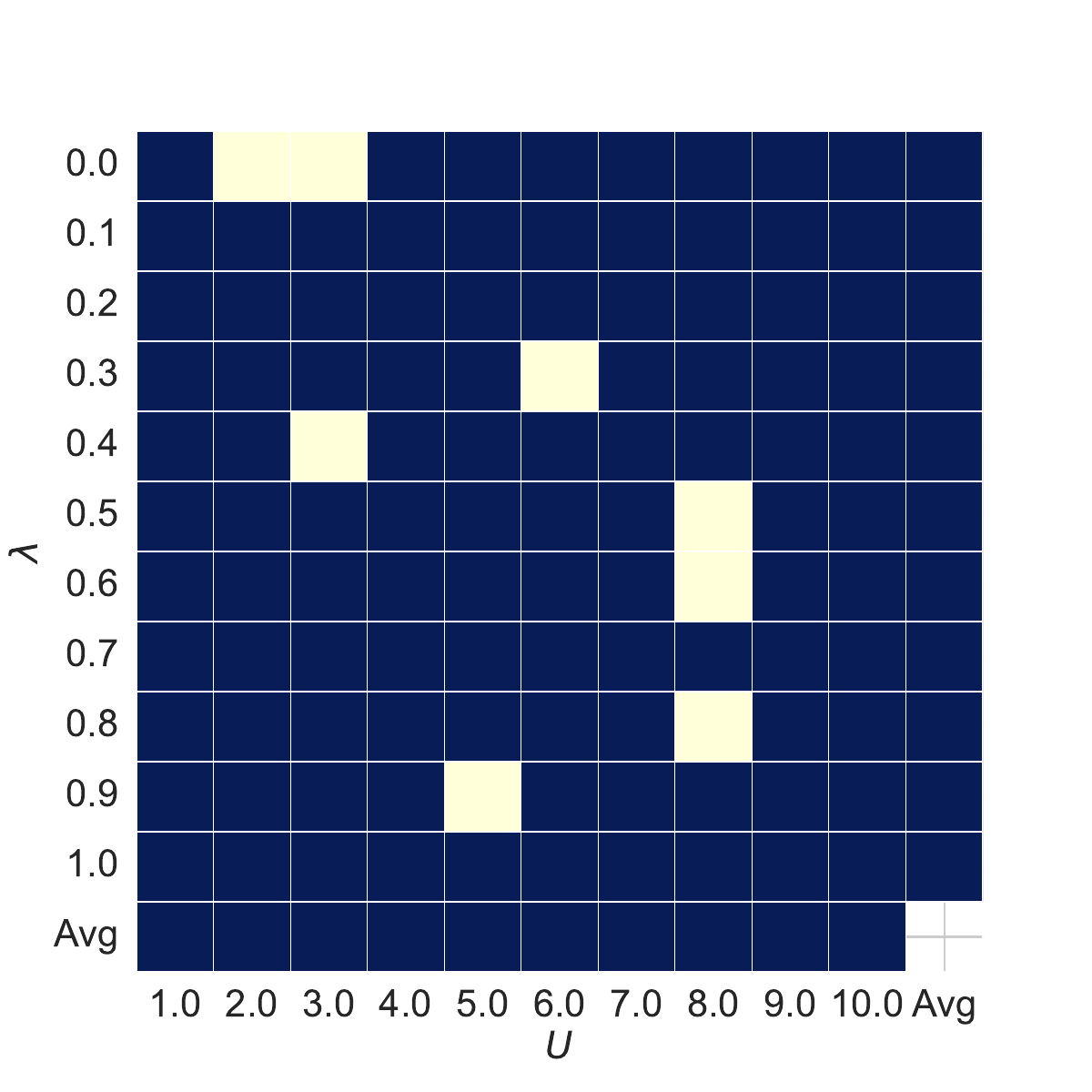}
      \caption{ACC w original features}
      \label{fig:sen_a}
  \end{subfigure}
  \hfill
  \begin{subfigure}[b]{0.24\textwidth}
      \centering
      \includegraphics[width=\textwidth]{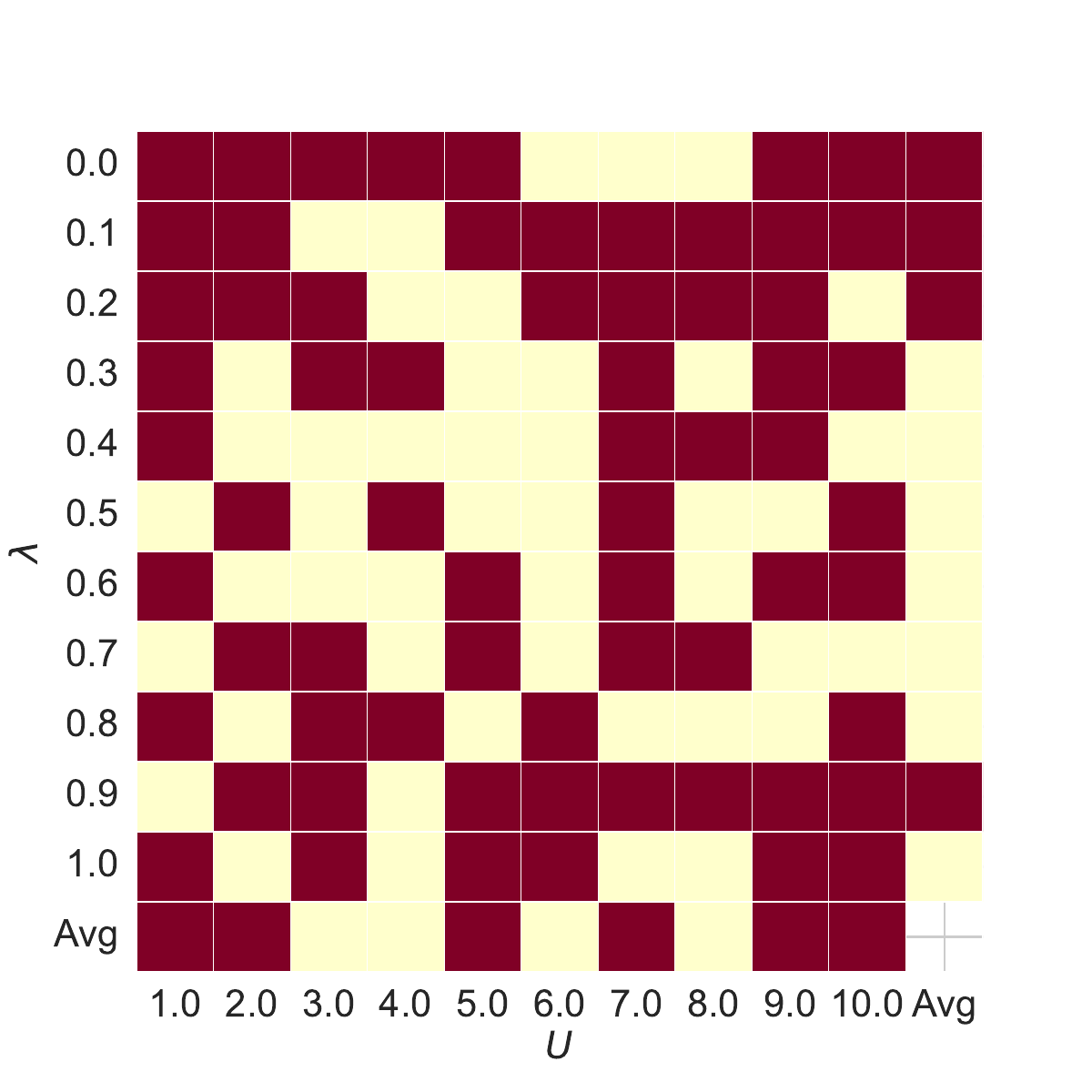}
      \caption{AUC w original features}
      \label{fig:sen_b}
  \end{subfigure}
  \hfill
  \begin{subfigure}[b]{0.24\textwidth}
      \centering
      \includegraphics[width=\textwidth]{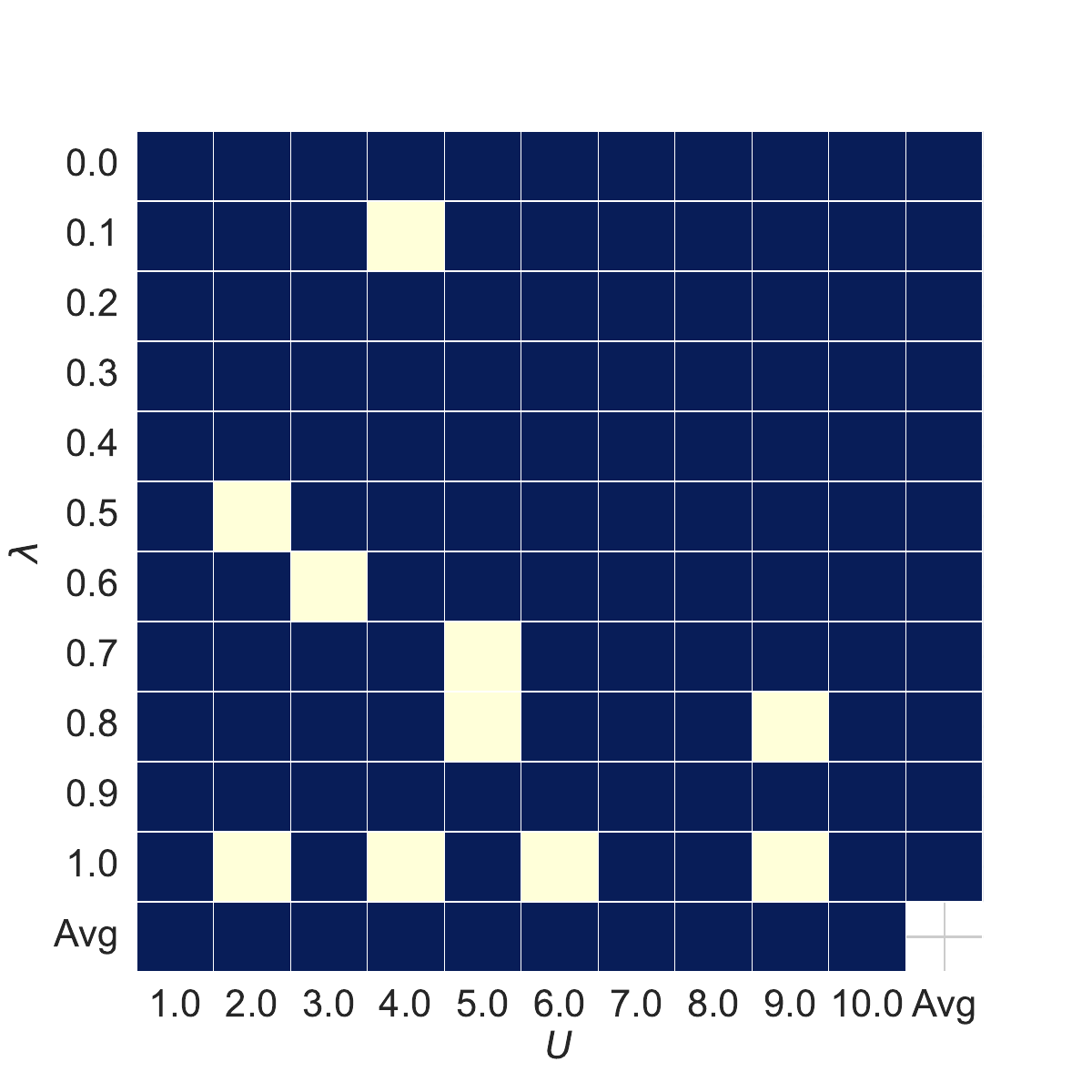}
      \caption{ACC w/o original features}
      \label{fig:sen_c}
  \end{subfigure}
  \hfill
  \begin{subfigure}[b]{0.24\textwidth}
      \centering
      \includegraphics[width=\textwidth]{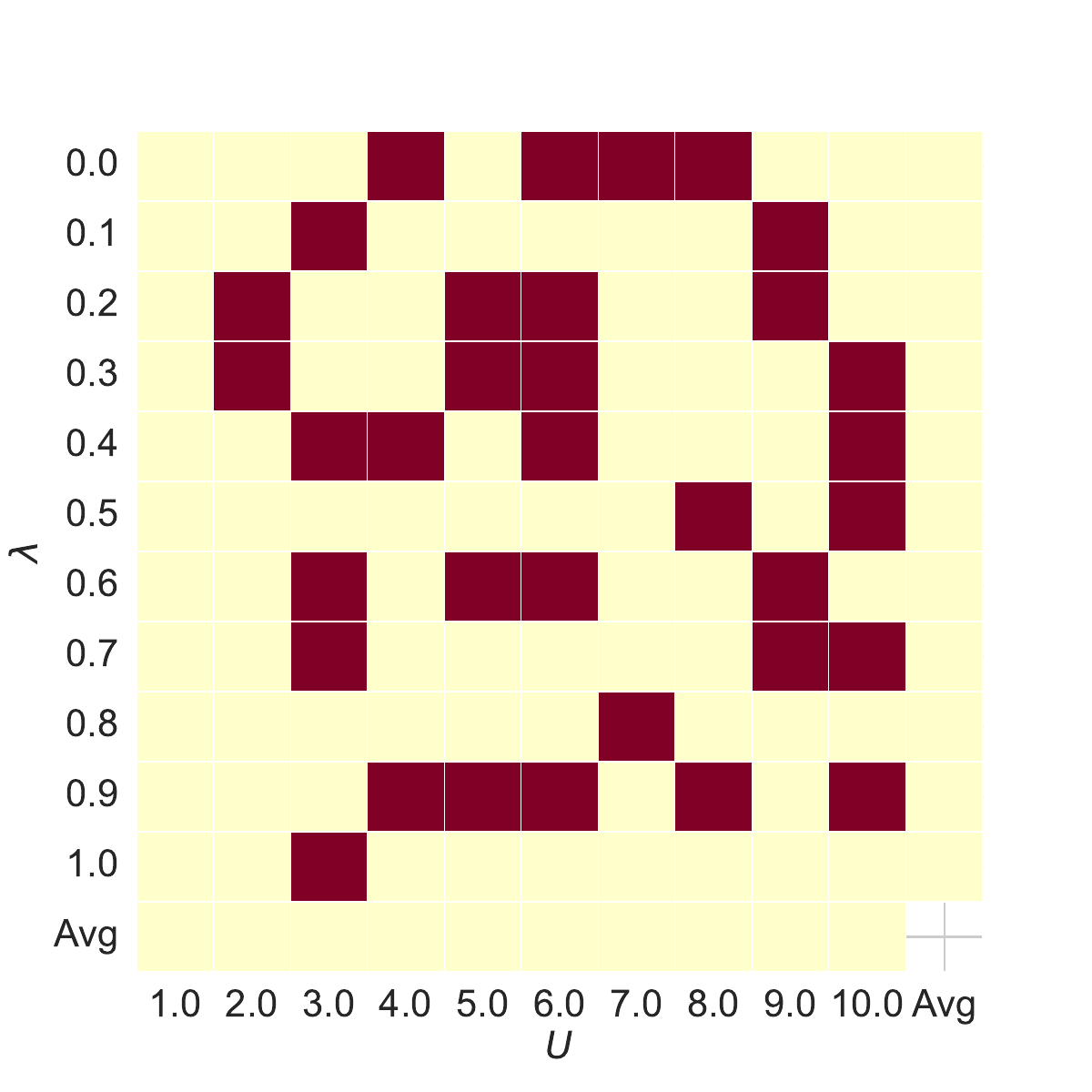}
      \caption{AUC w/o original features}
      \label{fig:sen_d}
  \end{subfigure}
  \caption{
    	 Ablation study for BSDA on BreastMNIST dataset using ResNet-18. 
  	 Under the same hyper-parameters, if BSDA improves performance relative to the baseline (AUC=$89.6\%$, ACC=$81.0\%$), the heatmap displays red(AUC) or blue(ACC).
  	 In each subplot, the horizontal axis represents the sampling rate $U$ of BSDA, while the vertical axis denotes the probability $\lambda$ of random direction selection. 
  	 ~\ref{fig:sen_a} and ~\ref{fig:sen_b} adds original features relative to ~\ref{fig:sen_c} and ~\ref{fig:sen_d}.
  }
  \label{fig: sensitivity_hyperparameters}
\end{figure}

\textbf{Hyper-parameters Analysis}
As shown in Figure~\ref{fig: sensitivity_hyperparameters}, subplots~\ref{fig:sen_a} and~\ref{fig:sen_b} show the AUC and ACC results of the joint features for different hyperparameters $\lambda$ and $U$ compared to the baseline, with red or blue indicating improvements over the baseline.
Comparing subplots~\ref{fig:sen_a} and~\ref{fig:sen_c} and~\ref{fig:sen_b} and~\ref{fig:sen_d}, we see that the joint features improve model performance in most cases.

\begin{figure}[t]
	\includegraphics[width=0.45\textwidth]{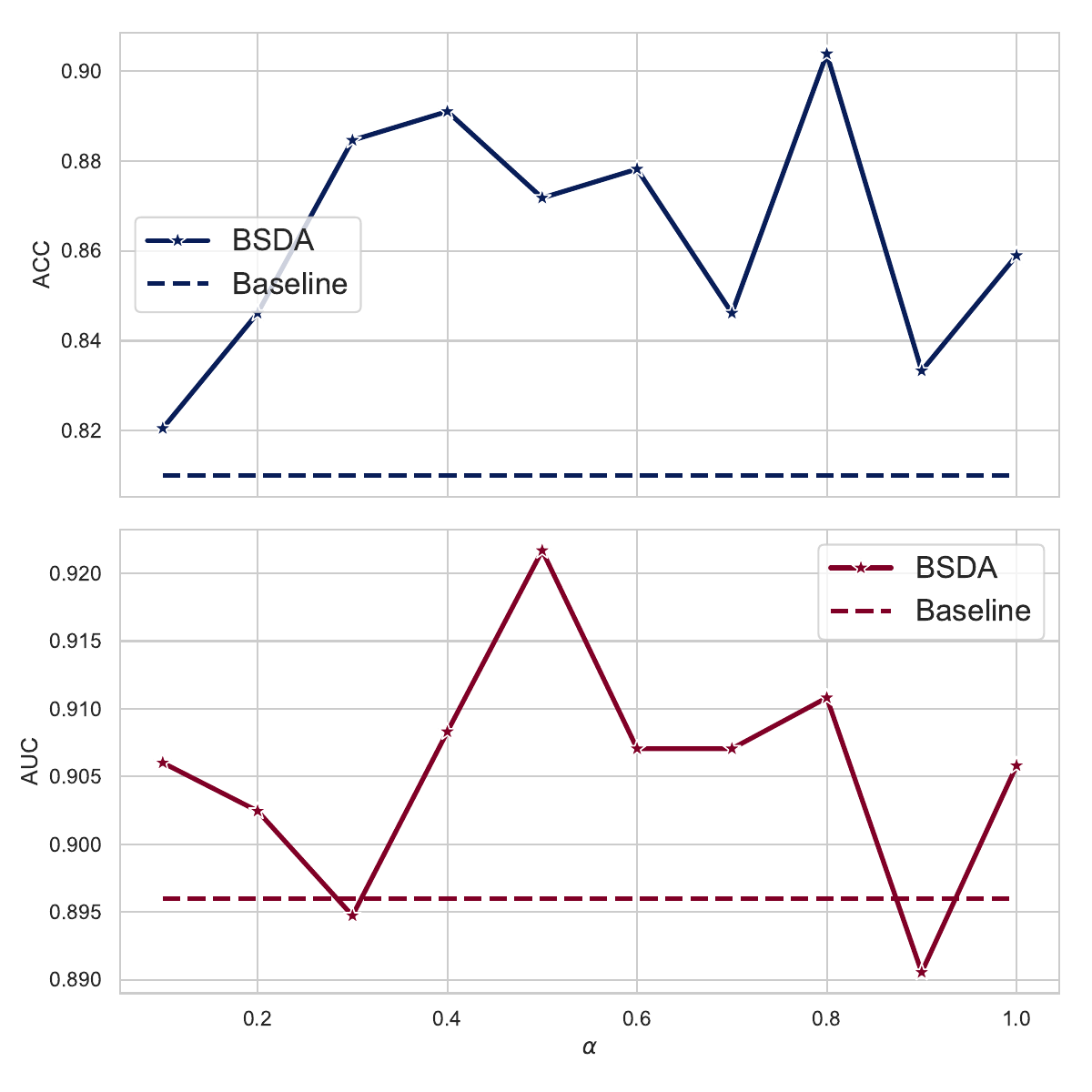}
	\caption{
        Sensitivity analysis was conducted on the parameter $\alpha$ of BSDA using ResNet-18 on the BreastMNIST dataset. 
        The horizontal axis indicates the values of $\alpha$, while the vertical axis illustrates the AUC and ACC performance metrics.		
        }
		\label{figures: sensitivity2}
\end{figure}

\textbf{Weight of Loss}
We conducted a sensitivity analysis of the hyperparameter $\alpha$ of BSDA using ResNet-18 on the BreastMNIST dataset.
The results are shown in Figure~\ref{figures: sensitivity2}.
We chose the hyperparameter $\alpha$ values from 0.1 to 1.
It can be observed that the AUC and ACC metrics are stable in almost all cases, indicating that the BSDA method is not sensitive to the hyperparameter $\alpha$.
The AUC value is slightly lower than the baseline at $\alpha = 0.3$ or $\alpha = 0.9$, and the ACC value is somewhat lower than the baseline at $\alpha = 0.1$ or $\alpha = 0.9$.
Empirically, we recommend $\alpha=0.5$ as the default parameter.

\subsection{Visualization of Deep Features}

\begin{figure}[t]
  \centering
	\includegraphics[width=0.45\textwidth]{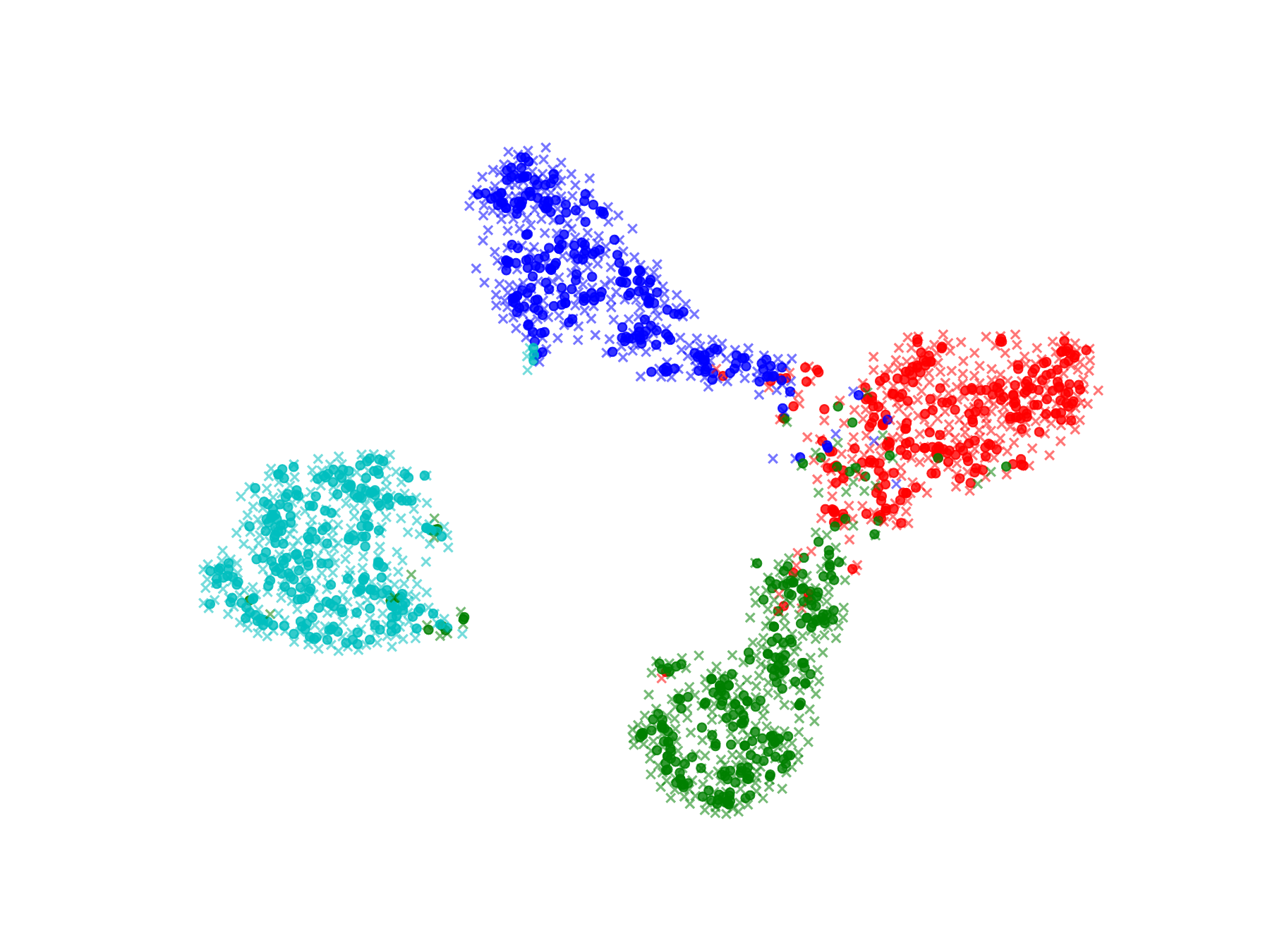}
	\caption{
		Visualizing deep features using t-SNE on OCTMNIST, individual colors correspond to specific categories. 
		Circular markers represent original features, while cross markers indicate augmented features generated by BSDA.
		}
		\label{fig: vis_tsne}
\end{figure}

We visualize the deep features using t-SNE~\cite{van2008visualizing} in Figure~\ref{fig: vis_tsne}.
Circular markers represent the original features, while the cross markers indicate the augmented features generated by BSDA.
The results show that the augmented features are distributed around the original features, 
indicating that BSDA generates augmented features that are close to the original ones.
Figure~\ref{fig: vis_tsne} also shows that the original features wrap tightly around the newly generated features. 
Intuitively, classifiers learned from the augmented features will be farther from the original features.

\begin{figure*}[!h]
  \centering
  \begin{subfigure}[b]{0.32\textwidth}
      \centering
      \includegraphics[width=\textwidth]{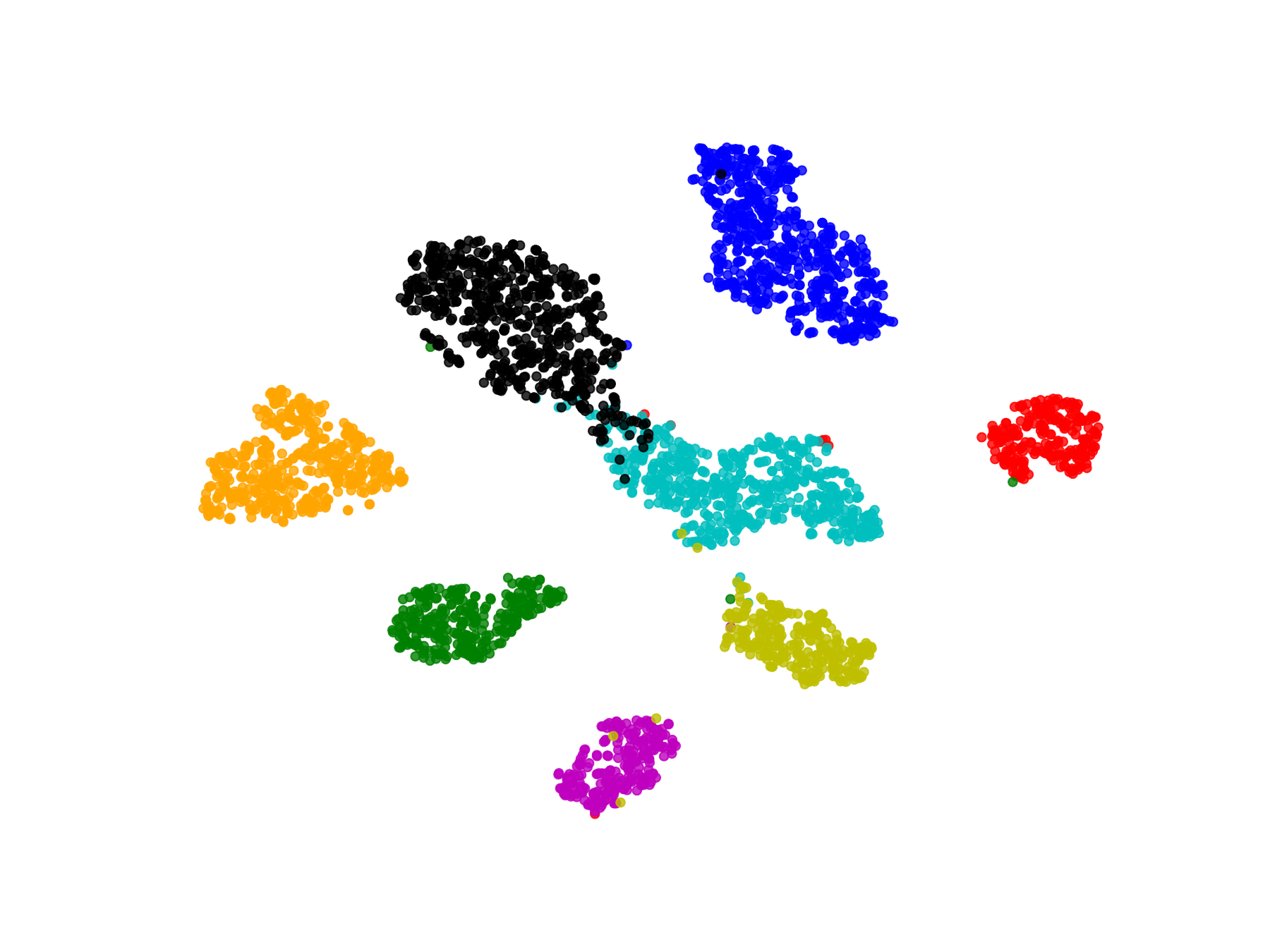}
      \caption{Baseline}
      \label{fig: vis_blood_baseline}
  \end{subfigure}
  \hfill
  \begin{subfigure}[b]{0.32\textwidth}
      \centering
      \includegraphics[width=\textwidth]{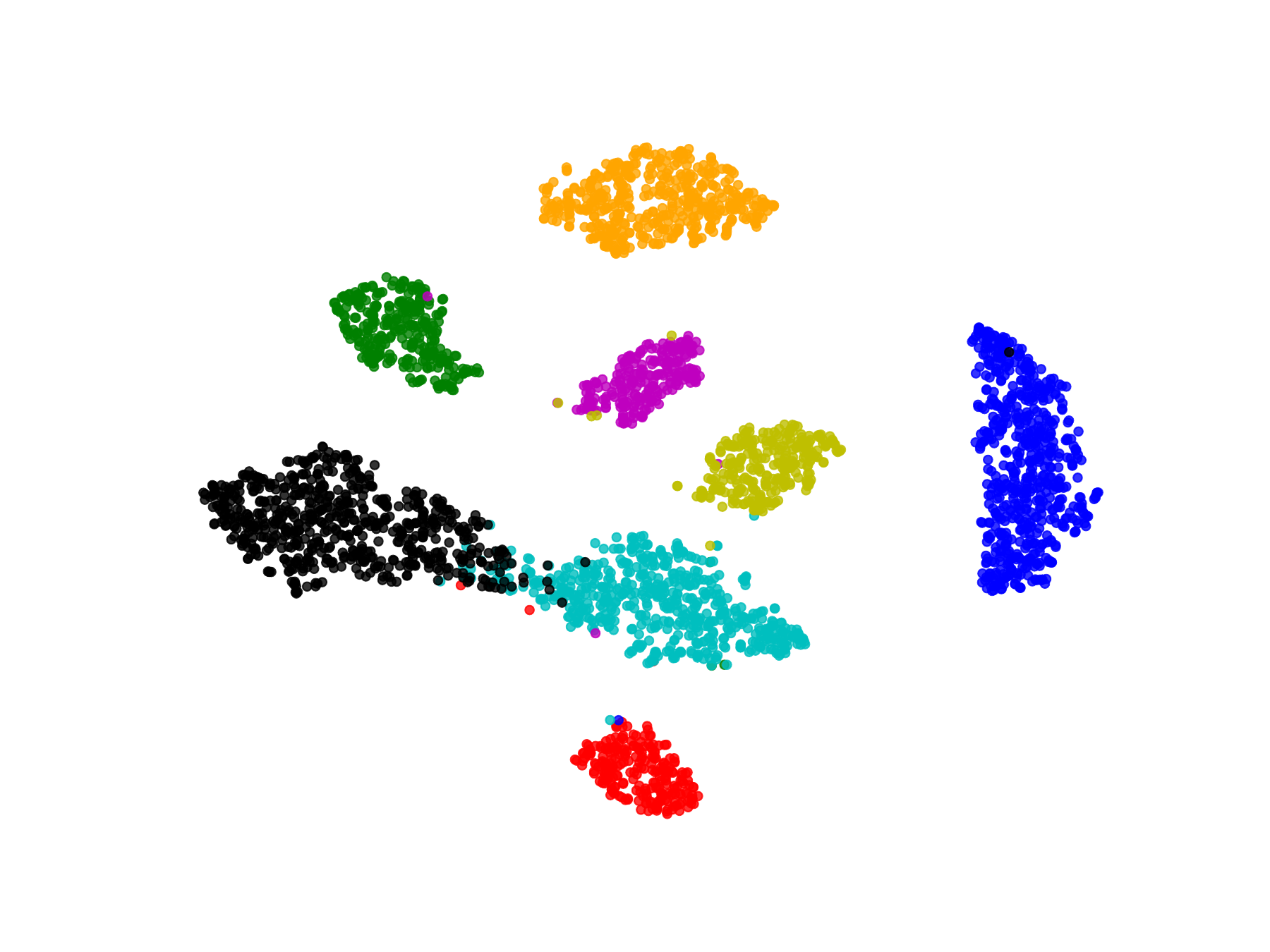}
      \caption{Mixup}
      \label{fig: vis_blood_mixup}
  \end{subfigure}
  \hfill
  \begin{subfigure}[b]{0.32\textwidth}
      \centering
      \includegraphics[width=\textwidth]{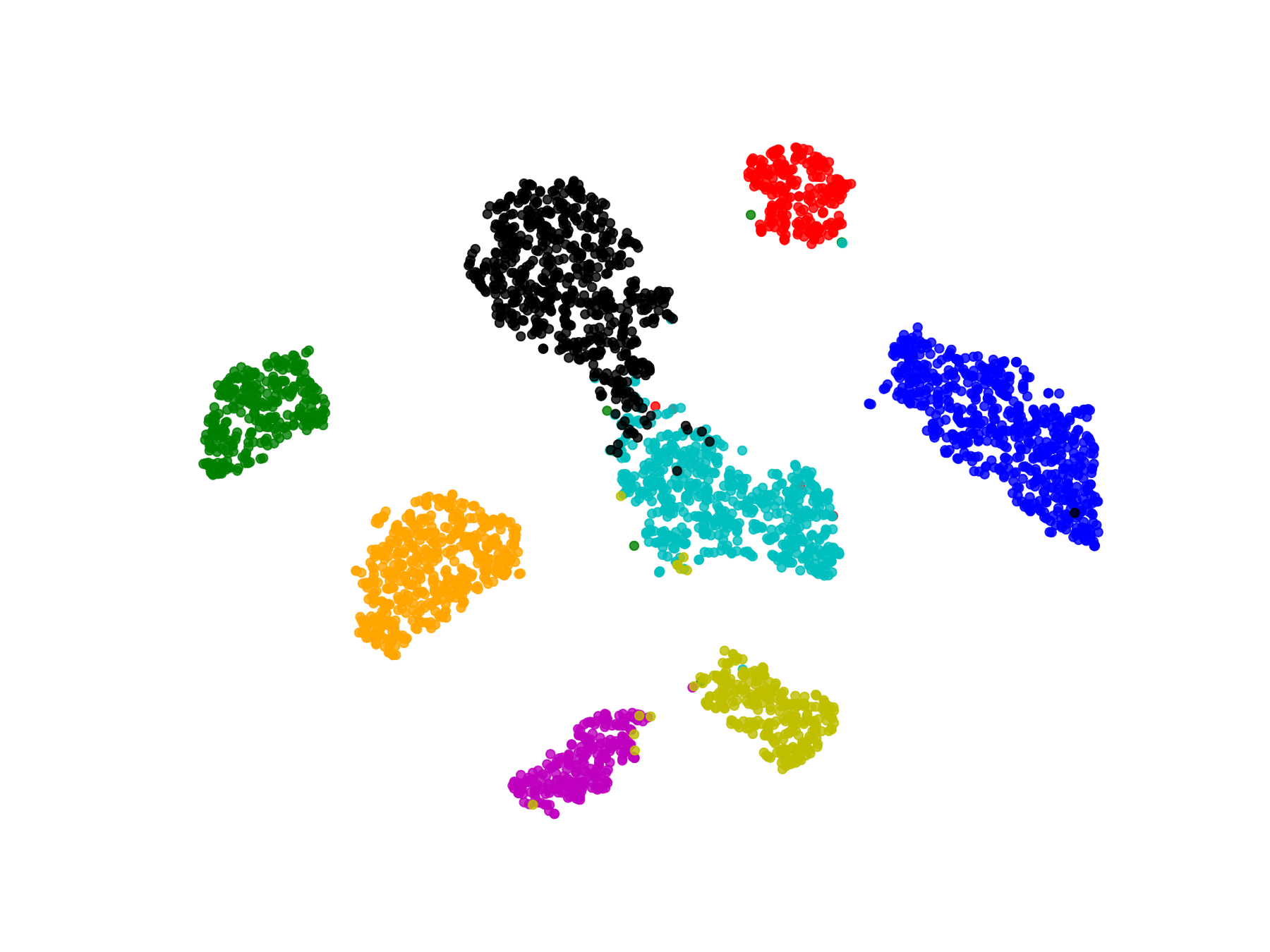}
      \caption{Cutout}
      \label{fig: vis_blood_cutout}
  \end{subfigure}
  \vfill
  \begin{subfigure}[b]{0.32\textwidth}
      \centering
      \includegraphics[width=\textwidth]{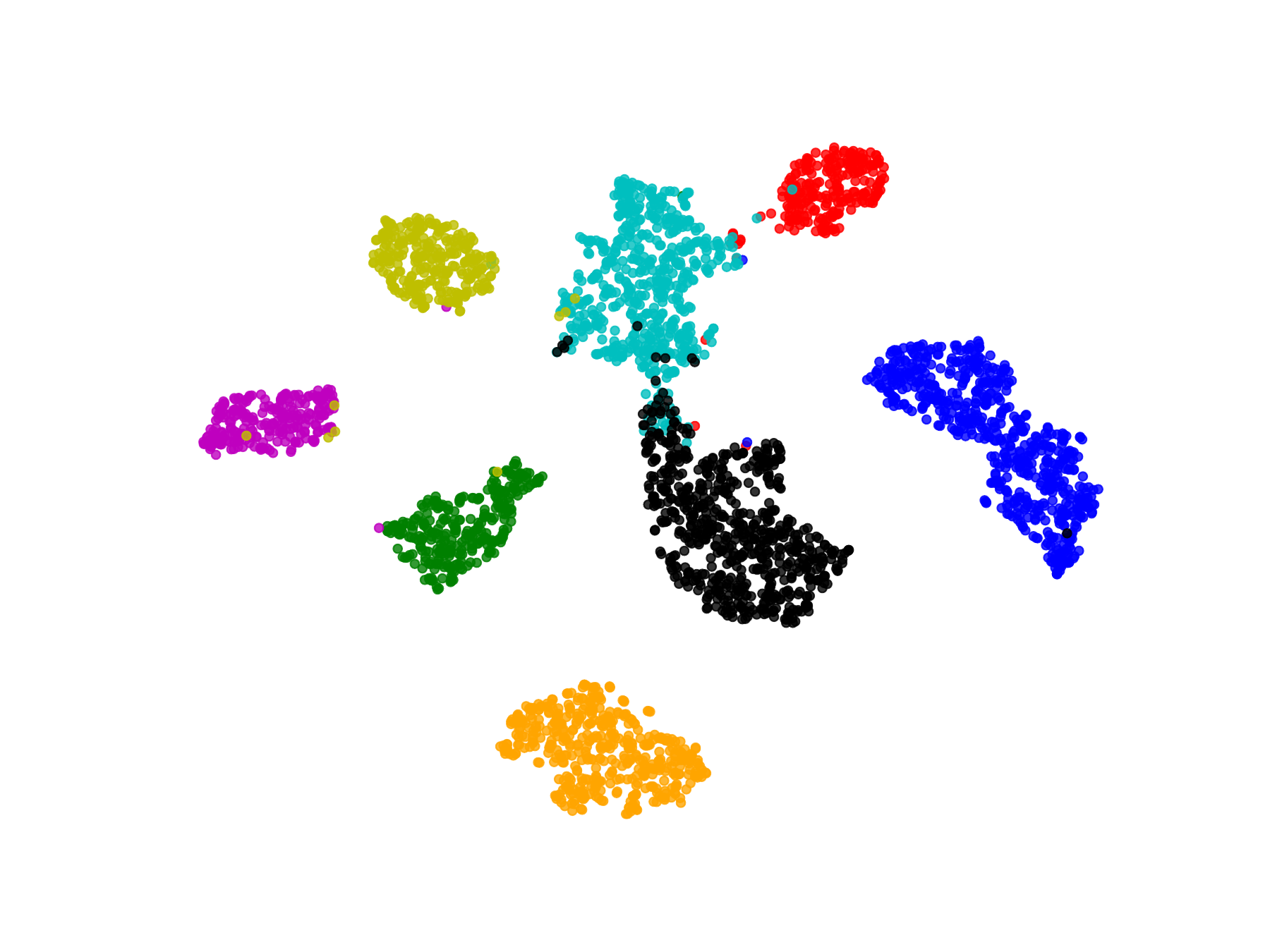}
      \caption{CutMix}
      \label{fig: vis_blood_cutmix}
  \end{subfigure}
  \hfill
  \begin{subfigure}[b]{0.32\textwidth}
      \centering
      \includegraphics[width=\textwidth]{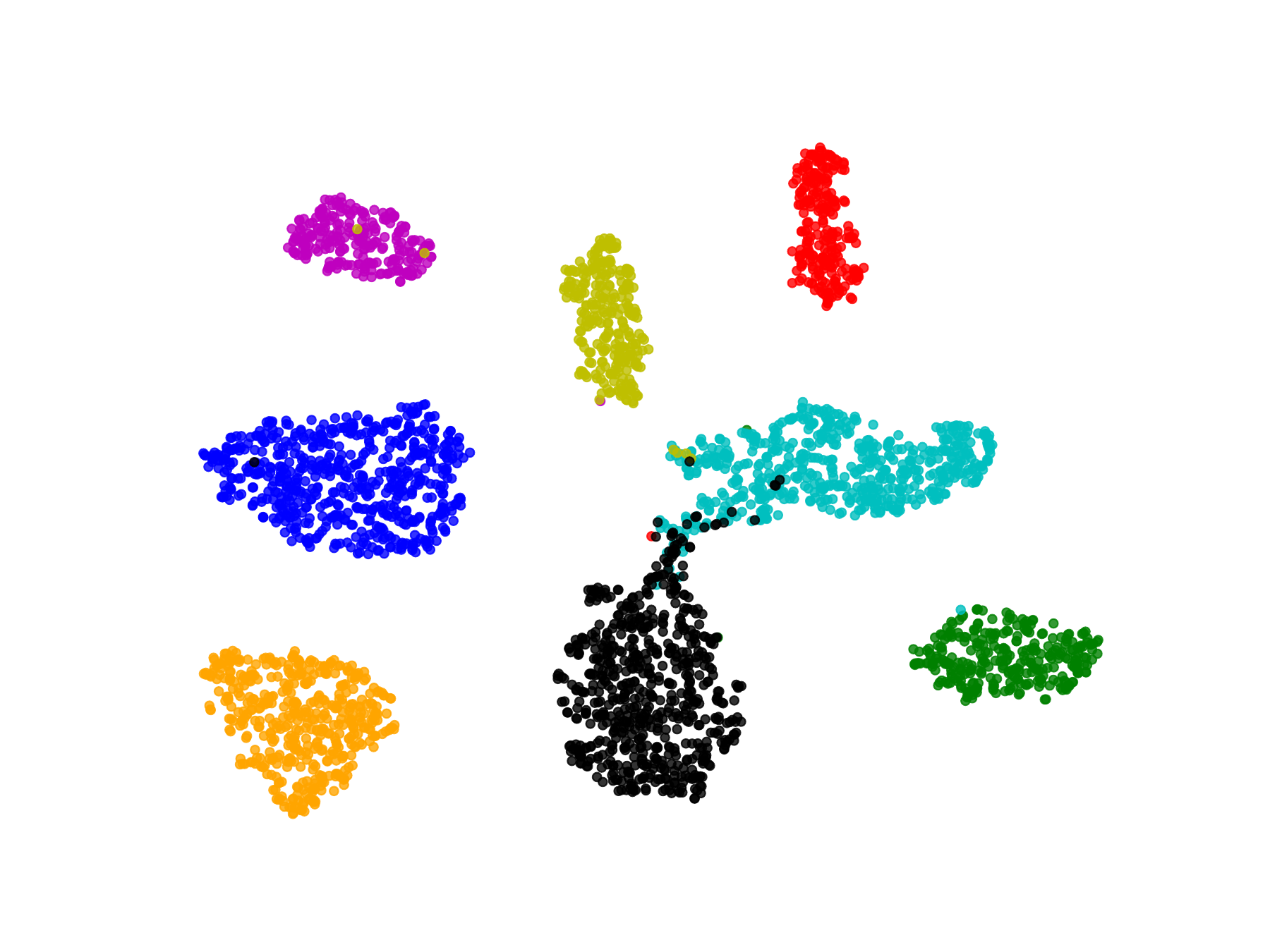}
      \caption{ISDA}
      \label{fig: vis_blood_isda}
  \end{subfigure}
  \hfill
  \begin{subfigure}[b]{0.32\textwidth}
      \centering
      \includegraphics[width=\textwidth]{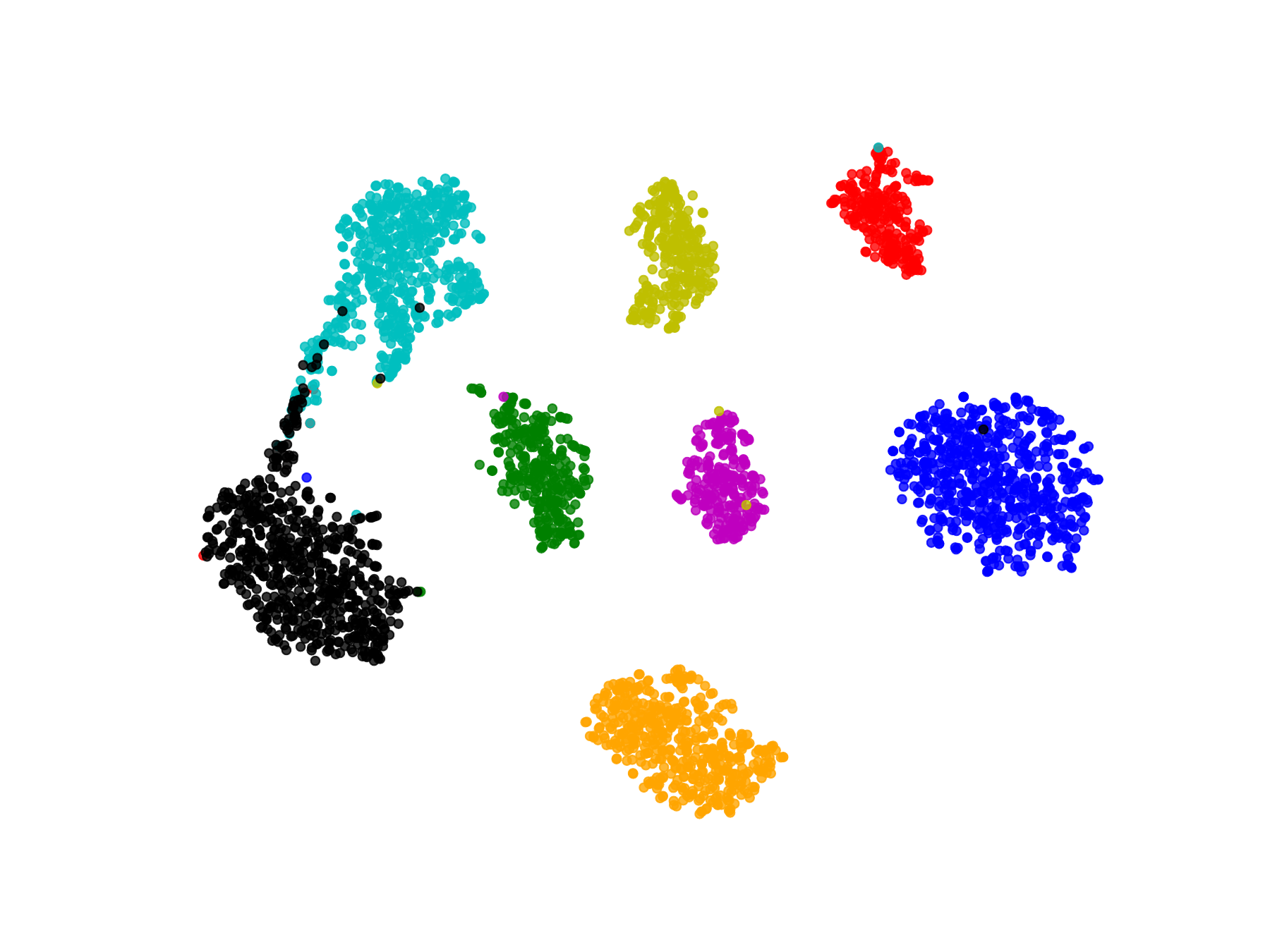}
      \caption{BSDA}
      \label{fig: vis_blood_bsda}
  \end{subfigure}
  \caption{
    Visualizing deep features using t-SNE on BloodMNIST, individual colors correspond to specific categories.
    BSDA methods (Fig.~\ref{fig: vis_blood_bsda}) make the intra-class features more cohesive and more easily separable.
        }
  \label{fig: vis_tsne_method}
\end{figure*}

We also visualize the features learned by our method compared to other methods (without feature-level or image-level augmentation).
Figure~\ref{fig: vis_tsne_method} shows that BSDA methods (Fig.~\ref{fig: vis_blood_bsda}) make the intra-class features more cohesive and more easily separable.

%% file: context/dis.tex
\section{Discussion}
BSDA, as an explicit data augmentation method, has many promising applications in medical imaging. 
For example, in the case of multimodal or multiparametric medical images, BSDA can be flexibly inserted into the network as a plug-in to provide semantic data augmentation for different modalities simultaneously, potentially enhancing the model's performance. 
Another example is the multicenter setting, typical in medical imaging, where the model must be generalized to other domains. 
Some methods use feature decoupling-based or feature disentanglement-based domain generalization. 
BSDA can also provide semantic data augmentation for different decoupled branches, potentially enhancing the model's generalization ability. 
Moreover, medical image datasets often need to be more balanced. Adding a balanced sampling strategy during data augmentation in the training phase can improve network performance on unbalanced datasets.

BSDA presents a versatile and efficient approach to data augmentation in medical imaging, promising significant improvements in model performance and generalization. 
Future work will optimize BSDA for specific medical imaging tasks and explore its potential in real-world clinical settings.

\section{Conclusion}
This paper introduces an efficient, plug-and-play Bayesian Random Semantic Data Augmentation (BSDA) method for medical image classification. 
BSDA generates new samples in feature space, making it more efficient and easy to implement. 
We experimentally demonstrate the effectiveness and efficiency of BSDA on various modalities, dimensional datasets, and networks.